\documentclass{article} 
\usepackage{iclr2026_conference,times}

\usepackage[utf8]{inputenc} 
\usepackage[T1]{fontenc}    
\usepackage{hyperref}       
\usepackage{url}            
\usepackage{booktabs}       
\usepackage{amsfonts}       
\usepackage{nicefrac}       
\usepackage{microtype}      
\usepackage{xcolor}         

\usepackage{graphicx}
\usepackage{multirow} 
\usepackage{booktabs}
\usepackage{colortbl}
\definecolor{gg}{HTML}{e2f0cb}
\usepackage{amssymb}
\usepackage{pifont}

\usepackage[most]{tcolorbox}
\tcbuselibrary{listings, breakable, skins}
\usepackage[dvipsnames]{xcolor}

\definecolor{mybrown}{rgb}{0.68627,0.55294,0.1459}
\definecolor{mypink}{rgb}{0.94901,0.4745,0.4431}

\newcommand{\cmark}{\ding{51}}

\usepackage{amsmath} 

\usepackage{hyperref}
\usepackage{url}

\iclrfinalcopy

\usepackage{caption}

\title{\raisebox{-0.85ex}{\includegraphics[width=1.5em]{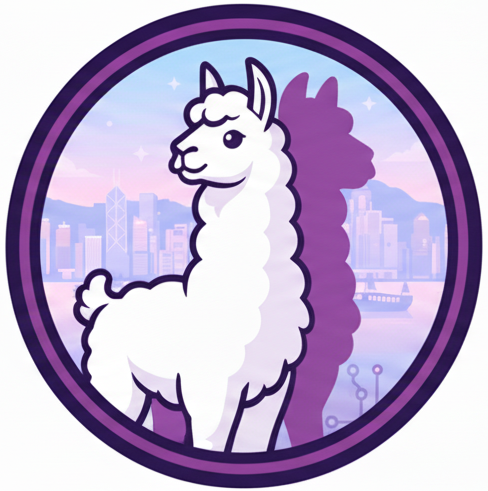}}\hspace{0.3em}Shadow-FT: Tuning Instruct Model via \\ Training on Paired Base Model}

\author{%
 Taiqiang Wu\textsuperscript{\rm 1}$^*$ \quad
 Runming Yang\textsuperscript{\rm 1}\thanks{Equal contributions. Work was done when Runming was interning at Tencent.} \quad \\
 \textbf{Jiayi Li}\textsuperscript{\rm 2} \quad
 \textbf{Pengfei Hu}\textsuperscript{\rm 3} \quad
 \textbf{Yik-Chung Wu}\textsuperscript{\rm 1} \quad
 \textbf{Ngai Wong}\textsuperscript{\rm 1} \quad
 \textbf{Yujiu Yang}\textsuperscript{\rm 2}\thanks{Corresponding author.} \quad
 \\
 \textsuperscript{\rm 1}The University of Hong Kong  \quad
 \textsuperscript{\rm 2}Tsinghua University
 \quad 
 \textsuperscript{\rm 3}Tencent\\
 \url{https://github.com/wutaiqiang/Shadow-FT} \\
}

\usepackage{makecell} 
\usepackage[table]{xcolor}
\usepackage{array}
\definecolor{gglight}{HTML}{FFFFFF}  
\definecolor{ggdark}{HTML}{e2f0cb}  %
\definecolor{ggtwo}{HTML}{F2F2F2}   
\newcolumntype{I}{>{\columncolor{gglight}}c}       
\newcolumntype{J}{>{\columncolor{ggdark}}c}   
\newcolumntype{B}{>{\columncolor{ggtwo}}c}    
\newlength{\seriesgap}
\newlength{\halfgap}
\newcolumntype{S}{>{\columncolor{ggdark}}c}  
\usepackage{wrapfig}      
\usepackage{graphicx}     
\usepackage{caption}      
\newcommand{\gut}[2]{%
  \begingroup
  \setlength{\fboxsep}{2pt}
  \colorbox{#1}{\strut #2}%
  \endgroup
}

\usepackage{booktabs}
\usepackage{multirow}
\usepackage[table]{xcolor} 
\usepackage{caption}

\usepackage{booktabs}

\begin{document}

\maketitle

\begin{abstract}
Large language models~(LLMs) consistently benefit from further fine-tuning on various tasks.
However, we observe that directly tuning the \textsc{Instruct}~(i.e., instruction-tuned) models often leads to marginal improvements and even performance degeneration.
Notably, paired \textsc{Base} models, the foundation for these \textsc{Instruct} variants, contain highly similar weight values~(i.e., less than 2\% on average for Llama 3.1 8B).
The \textsc{Base} model tends to be a good learner yet a weak backbone without post-training.
Therefore, we propose a novel \textbf{Shadow-FT} framework to tune the \textsc{Instruct} models by leveraging corresponding \textsc{Base} models.
The key insight is to fine-tune the \textsc{Base} model, and then \textit{directly} graft the learned weight updates to the \textsc{Instruct} model.
Our proposed Shadow-FT introduces no additional parameters, is easy to implement, and significantly improves performance.
We conduct extensive experiments on tuning mainstream LLMs, such as Qwen 3 and Llama 3 series, and evaluate them across 19 benchmarks covering coding, reasoning, and mathematical tasks.
Experimental results demonstrate that Shadow-FT consistently outperforms conventional full-parameter and parameter-efficient tuning approaches.
Further analyses indicate that Shadow-FT can be applied to multimodal large language models~(MLLMs) and combined with direct preference optimization~(DPO).
\end{abstract}

\section{Introduction}

Large Language Models~(LLMs), such as Qwen~\citep{bai2023qwen}, Llama~\citep{llama3modelcard}, and Gemma~\citep{team2025gemma}, have demonstrated remarkable performance across diverse disciplines~\citep{zhang2023instruction, wang2024large}.
Such a strong capability is always attributed to the pre-training on massive data with billions of parameters~\citep{bi2024deepseek, tao2024scaling}.
When applied in real-world scenarios, there are several challenges.
The users want the LLMs to follow their instructions helpfully and honestly~\citep{li2024survey}, which is not covered during the pre-training~\citep{zhang2023instruction, liu2024tuning}.
Meanwhile, the downstream tasks always involve specific domain knowledge requiring adaptation~\citep{wang2023huatuo,luo2024taiyi}.

To tackle these issues, one predominant approach is further tuning LLMs on desired tasks, including full parameter fine-tuning and parameter-efficient fine-tuning~\citep{liu2021p, hu2022lora}.
Typically, for each model size, two paired variants are provided: the pretrained base model (denoted as \textsc{Base}) and its instruction-tuned version (denoted as \textsc{Instruct}).
The \textsc{Base} model exhibits relatively poor instruction-following ability~(i.e., \textit{a weak backbone}), while the \textsc{Instruct} model performs better.
However, we observe that tuning the \textsc{Instruct} models brings marginal improvements and even a performance degeneration.
Therefore, how to tune the \textsc{Instruct} model effectively gains increasing importance.

\begin{figure*}[t!]
\centering
\includegraphics[width=\textwidth]{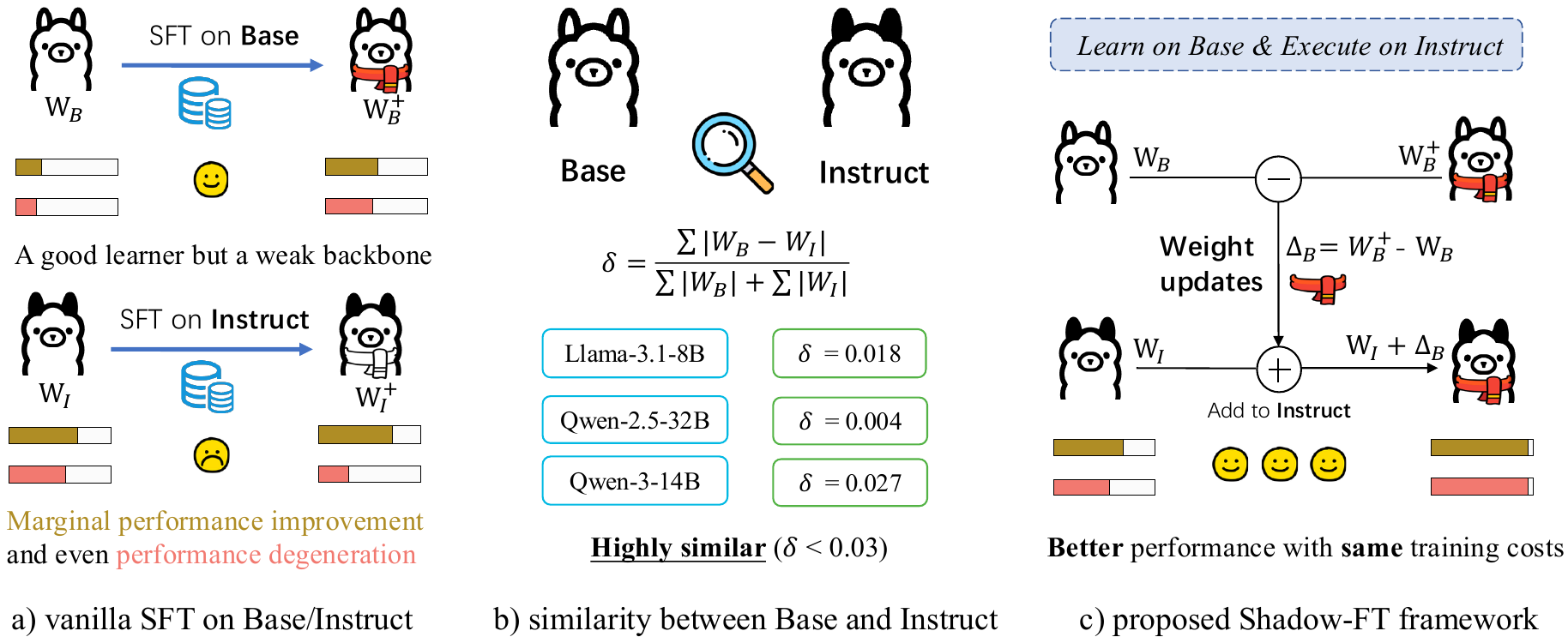}
\caption{
Performance of vanilla SFT (part a), similarity on weights(part b), and the Shadow-FT framework (part c).
The progress bars in \textcolor{mybrown}{brown} and \textcolor{mypink}{pink} denote the different abilities, \textbf{the fuller, the better}.
Based on the SFT dynamics and weight similarity (gap $\sigma$ less than 0.03), we propose to tune the paired \textsc{Base} model and then graft the weight updates onto \textsc{Instruct} model.
}
\label{fig: overview}
\end{figure*}

In this paper, we first analyze the weights of paired \textsc{Base} and \textsc{Instruct} models considering the relative absolute difference $\sigma$.
Fortunately, we find that the weights of \textsc{Base} and \textsc{Instruct} are highly similar.
As shown in Figure \ref{fig: overview}, the gap $\sigma$ is quite low, with an average $\sigma$ of 0.016 for the Llama-3.1-8B model.
Intuitively, the contained instruction-following ability of \textsc{Instruct} model disturbs the learning of new knowledge, while \textsc{Base} can avoid it.
We further provide a deep analysis to prove this conclusion.
Motivated by these, we thus propose a novel \textbf{Shadow-FT} framework to employ the \textsc{Base} model as 'shadow' of \textsc{Instruct}.
The key is to tune the \textsc{Base} for better weight updates and directly graft these updates to \textsc{Instruct}, as they share the same structures.

To evaluate the performance, we conduct extensive experiments tuning mainstream LLMs such as Qwen 3 \citep{bai2023qwen} and Llama 3~\citep{llama3modelcard}.
For the tuning data, we employ the BAAI-Infinity-Instruct Dataset\footnote{https://huggingface.co/datasets/BAAI/Infinity-Instruct} and extract 2000 samples named as BAAI-2k following~\citep{zhou2023lima, muennighoff2025s1}.
Without the loss of generality, we apply Shadow-FT on full parameter and low-rank settings, and then report the performance on 19 datasets.
Experimental results indicate that Shadow-FT consistently outperforms the baselines under various settings, demonstrating its effectiveness and robustness.
Further analyses show that Shadow-FT can be applied to MLLMs and combined with DPO for alignment.
Our contributions can be concluded as follows:
\begin{itemize}
\item We find that paired \textsc{Base} and \textsc{Instruct} are highly similar considering weight values, and thus propose a novel Shadow-FT framework.
The key is to tune the \textsc{Base} for better weight updates and directly graft these updates to \textsc{Instruct}.

\item We conduct extensive experiments tuning various mainstream LLMs and report the performance on 19 benchmarks across math, code, and reasoning.
Experimental results demonstrate the effectiveness and robustness of Shadow-FT.

\item This work highlights the potential of leveraging \textsc{Base} models to enhance their \textsc{Instruct} counterparts, and we hope it inspires further research and broader applications in the future.
\end{itemize}
\section{Preliminaries and Motivation}

\subsection{Background}

\paragraph{Basic tuning methods.}
Supervised Fine-tuning~(SFT) is a fundamental approach to updating the knowledge of LLMs.
Vanilla SFT methods update all the parameters by gradient descent following $W^+ \leftarrow W+\Delta W$, where $W \in \mathbb{R}^{d_1 \times d_2}$ is an arbitrary weight and $W^+$ is the updated variant.
To reduce the update costs, LoRA \citep{hu2022lora} introduces a low-rank branch to learn the weight updates following $W^+ \leftarrow W+AB$, where $A \in \mathbb{R}^{d_1 \times r}, B \in \mathbb{R}^{r \times d_2}$ and $r \ll \min\{d_1,d_2\}$.
The original weight $W$ is frozen during training, and only the low-rank branch is updated.

\paragraph{\textsc{Base} and \textsc{Instruct}.}
Current LLMs typically follow a two-stage training pipeline, including pre-training and post-training.
During pre-training, LLMs are trained on massive training data on next token prediction tasks \citep{brown2020language}, and the weights would be released as \textsc{Base} version.
The \textsc{Instruct} variant, post-trained upon the \textsc{Base} model, is further aligned with human preference and tuned for reasoning tasks \citep{ouyang2022training}.
Therefore, \textsc{Instruct} model performs better than \textsc{Base} model regarding instruction-following ability.

\subsection{Directly Tuning \textsc{Instruct}}

However, tuning the \textsc{Instruct} models often leads to marginal improvements and even performance degeneration.
Table \ref{table: main_res} shows the tuned performance of the \textsc{Instruct} models using the BAAI-2k.
We report the average scores of popular benchmarks.
Compared to the vanilla \textsc{Instruct}, the tuned version shows marginal improvement, and even degeneration in more cases.
Specifically, as shown in Table \ref{table: main_res}, tuning Qwen-3-4B on the BAAI-2k dataset via conventional LoRA would lead to a drop of 2.6 in Math-7 (from 73.8 to 71.2), 6.8 in Code-3 (from 66.4 to 59.6), and 2.6 in Knowledge-9 (from 63.7 to 61.1).
Therefore, how to effectively tune \textsc{Instruct} remains a challenge.

\subsection{Similar Weights: \textsc{Base} \& \textsc{Instruct}}

\begin{figure*}[t!]
\centering
\includegraphics[width=\textwidth]{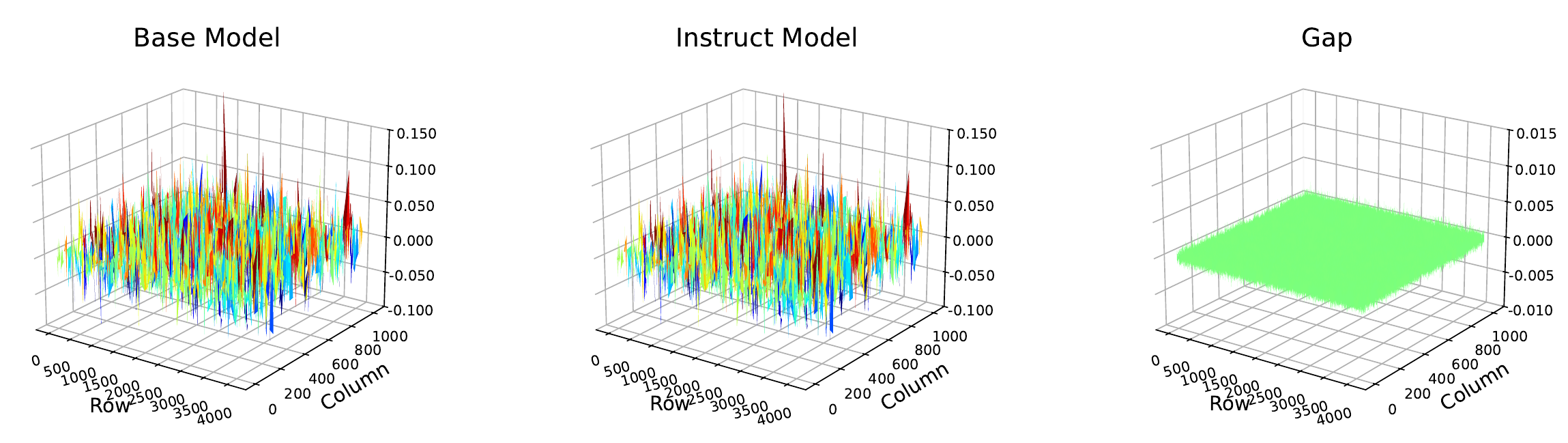}
\caption{
Weight distributions for Llama-3.1-8B.
We visualize the same linear layer~(layer.0.k\_proj) for \textsc{Base} model (left), \textsc{Instruct} model (middle), and their gap (right).
Though zoomed in 10x in the z-axis, the gap is negligible and the average $\sigma$ value is 0.016.
}
\label{fig: weight_diff}
\end{figure*}

Fortunately, we observe that the weights of \textsc{Base} and \textsc{Instruct} are highly similar.
To calculate the similarity, we first define the relative gap ratio $\sigma$ as follows:
\begin{equation}
    \sigma = \frac{\sum{|W_B-W_I|}}{\sum{|W_B|}+\sum{|W_I|}},
\end{equation}
where $\sum$ is the element-wise sum and $|\cdot|$ means the absolute operations.
The $\sigma$ would be 1 if one is much larger than the other, and be 0 if the two matrices are exactly the same. 
The smaller the $\sigma$, the more similar the two matrices are.
Figure \ref{fig: weight_diff} shows the weights of the same layer from \textsc{Base} and \textsc{Instruct}, and also their differences with $\sigma=0.016$.
We can find that the gaps are quite small and negligible after zooming in 10x in the z-axis.
Please refer to Appendix \ref{appendix: more_simi} for more $\sigma$ regarding various LLMs.
In summary, these paired \textsc{Base} and \textsc{Instruct} models are highly similar with $\sigma<0.03$.
\section{Methodology}

\subsection{Shadow-FT}

To tackle the issue that directly tuning \textsc{Instruct} fails, we propose a novel framework, Shadow-FT, to tune the \textsc{Instruct} on \textsc{Base}.
Motivated by the observation that \textsc{Base} and \textsc{Instruct} models are highly similar, we argue that the weight updates of \textsc{Base} can be directly added to \textsc{Instruct}.
Since they share the same structures, no extra operations are required.
Specifically, in Shadow-FT, we first tune the $\textsc{Base}$ model:
\begin{equation}
    W_B^{+} \leftarrow Tune(W_B),
\end{equation}
where $Tune$ is the fine-tuning method, such as full-parameter fine-tuning and LoRA.
After that, we would like to get the weights updates as the learned knowledge, and directly graft these updates to the $\textsc{Instruct}$ model as:
\begin{equation}
    W_{I}^{+} = W_I + (W_B^{+} - W_B) =W_I + (Tune(W_B) - W_B) .
\end{equation}
Traditional tuning on \textsc{Instruct} can be formulated as:
\begin{equation}
    W_{I}^{+} = W_I + (W_{I}^{+}-W_I) = W_I + (Tune(W_{I})-W_I). 
\end{equation}
We can find that Shadow-FT introduces no extra training costs. 
The only difference is the basic weights to learn the weight updates for \textsc{Instruct} model.
Vanilla FT methods rely on the \textsc{Instruct} model while Shadow-FT on the \textsc{Base} model.
Since the \textsc{Base} version is pre-trained only, we believe that the weight updates would be more suitable for modeling the knowledge with less priority, compared to updates of the \textsc{Instruct} version.

\subsection{Relation with Task Vectors}
Task Vectors aim to represent the ability on tasks as vectors, and are widely used for arithmetic operations on these tasks regarding the same base model \citep{ilharco2022editing}.
Chat Vector \citep{huang2023chat} extends such an idea to LLMs, which models weight differences between \textsc{Instruct} and \textsc{Base} models as vectors and then adds the vectors to continually pretrained \textsc{Base} models.
Specifically, Chat Vector continually pre-trains Llama2 \citep{touvron2023llama} on the Traditional Chinese corpus, and then adds on the chat vectors.
Compared to Chat Vector \citep{huang2023chat}, the differences are as follows:
1) \textit{task}: Chat Vector focuses on continual pertaining while Shadow-FT can be applied to board tuning methods, including full-parameter fine-tuning, LoRA, and DPO.
2) \textit{motivation}: Chat Vector aims to extend the language ability.
Shadow-FT aims to tackle the degeneration issue based on the weight similarity.

\section{Experiments}

\subsection{Experimental Setup}
\label{Exp:setting}

\paragraph{Training.}
For the tuning data, we build BAAI-2k by extracting 2000 samples from BAAI-Infinity-Instruct Dataset \footnote{https://huggingface.co/datasets/BAAI/Infinity-Instruct/tree/main/Gen} following~\citep{zhou2023lima, muennighoff2025s1}.
We select the samples with high rewards to ensure the data quality and uniform sampling among all categories for data diversity.
Without loss of generality, we tune various LLMs, including Qwen 3 series \citep{qwen3} and Llama 3 series \citep{llama3modelcard}.
Also, we report the results on Gemma-3 series \citep{team2025gemma}, Yi series \citep{young2024yi}, and Falcon series \citep{almazrouei2023falcon} in Section \ref{sec: zoo}.
We employ LLaMA-Factory \citep{zheng2024llamafactory} for the code base and apply two tuning strategies: full-parameter and LoRA.
All experiments are conducted on 8 A100 GPUs.
Please refer to Appendix \ref{appendix: hyper} for detailed hyperparameters.

\paragraph{Evaluation.}
To evaluate the tuned LLMs on downstream benchmarks, we employ the OpenCompass framework \citep{2023opencompass} and lmdeploy as the acceleration framework~\citep{2023lmdeploy}.
During inference, we set the cutoff length as 4096 and the batch size as 512.
Considering the benchmarks, we select three representative abilities, i.e., mathematical, coding, and commonsense reasoning ability, and report the average scores marked as Math-7, Code-3, and Knowledge-9.
Specifically, Math-7 denotes the results of AIME24~\cite{AIME2024}, GSM8K~(0-shot and 8-shot)~\cite{gsm8k}, MATH~\cite{hendrycksmath2021}, MATH-500, Minerva\_Math~\cite{2206.14858}, SVAMP~\cite{patel-etal-2021-nlp}.
Code-3 for HumanEval~\cite{chen2021codex}, HumanEval+~\cite{evalplus}, LiveCodeBench~\cite{livecodebench}.
Knowledge-9 for ARC-challenge~\cite{allenai:arc}, BBH~(0-shot and few-shot), DROP~\cite{Dua2019DROP}, GPQA Diamond~\cite{rein2024gpqa}, MMLU~\cite{hendryckstest2021}, MMLU Pro~\cite{wang2024mmlu}, Winogrande~\cite{ai2:winogrande}, TheoremQA~\cite{chen2023theoremqa}.
To avoid the impact of different prompts, we mainly evaluate under a zero-shot setting.
Please refer to Appendix \ref{appendix: benchmark} for more details.
For Qwen-3 series, we adapt enable\_thinking as false for universal evaluations,
and we report pass@k results of both thinking and non-thinking in Appendix~\ref{appendix: pass@k}.
\subsection{Main Results}

\begin{table}[!t]
\centering
\caption{Performance comparison of different methods tuning popular LLMs. \textbf{Math-7} denotes the average score of 7 mathematical benchmarks including AIME24, \textbf{Code-3} for 3 code benchmarks including LiveCodeBench, and \textbf{Knowledge-9} for 9 commonsense reasoning benchmarks including MMLU Pro.
For \textbf{Math-7} and \textbf{Code-3}, we report the mean value of three runs.
We employ the Instruct version and report the final average scores.
Please refer to Appendix \ref{appendix: detail_data} for detailed scores.
}
\renewcommand{\arraystretch}{1.1}
\label{table: main_res}
\resizebox{0.95\textwidth}{!}{

\begin{tabular}{ccccccccc}
\toprule
\multirow{2}{*}{\textbf{Model}} & \multirow{2}{*}{\textbf{Method}} & \multicolumn{2}{c}{\textbf{Math-7}} & \multicolumn{2}{c}{\textbf{Code-3}} & \multicolumn{2}{c}{\textbf{Knowledge-9}} & \multirow{2}{*}{\textbf{Avg.}} \\
\cmidrule(lr){3-4} \cmidrule(lr){5-6} \cmidrule(lr){7-8}
 &  & \textit{Full} & \textit{LoRA} & \textit{Full} & \textit{LoRA} & \textit{Full} & \textit{LoRA} \\
\midrule

\multirow{3}{*}{Qwen-3-4B} & Instruct & \multicolumn{2}{c}{73.8} & \multicolumn{2}{c}{66.4} & \multicolumn{2}{c}{63.7} & 68.0 \\
\cmidrule(lr){2-2} \cmidrule(lr){3-4} \cmidrule(lr){5-6} \cmidrule(lr){7-8} \cmidrule(lr){9-9}
 & FT & 72.9 & 71.2 & 66.4 & 59.6 & 62.9 & 61.1 & 65.7 \\
 & \cellcolor{gg}{Shadow-FT} & 73.7 & 75.9 & 67.4 & 69.7 & 64.9 & 65.0 & \cellcolor{gg}{\textbf{69.4}} \\
\midrule
\multirow{3}{*}{Qwen-3-8B} & Instruct & \multicolumn{2}{c}{74.5} & \multicolumn{2}{c}{72.7} & \multicolumn{2}{c}{64.7} & 70.6 \\
\cmidrule(lr){2-2} \cmidrule(lr){3-4} \cmidrule(lr){5-6} \cmidrule(lr){7-8} \cmidrule(lr){9-9}
 & FT & 74.0 & 71.3 & 71.2 & 69.6 & 64.6 & 64.3 & 69.2 \\
 & \cellcolor{gg}{Shadow-FT} & 75.9 & 74.8 & 73.1 & 71.9 & 65.6 & 67.8 & \cellcolor{gg}{\textbf{71.5}} \\
\midrule
\multirow{3}{*}{Qwen-3-14B} & Instruct & \multicolumn{2}{c}{75.8} & \multicolumn{2}{c}{76.8} & \multicolumn{2}{c}{71.2} & 74.6 \\
\cmidrule(lr){2-2} \cmidrule(lr){3-4} \cmidrule(lr){5-6} \cmidrule(lr){7-8} \cmidrule(lr){9-9}
 & FT & 75.2 & 73.3 & 76.2 & 74.4 & 70.6 & 70.4 & 73.4 \\
 & \cellcolor{gg}{Shadow-FT} & 78.9 & 78.6 & 77.0 & 77.8 & 71.4 & 71.5 & \cellcolor{gg}{\textbf{75.9}} \\
\midrule
\multirow{3}{*}{Qwen-2.5-32B} & Instruct & \multicolumn{2}{c}{74.1} & \multicolumn{2}{c}{75.9} & \multicolumn{2}{c}{73.4} & 74.5 \\
\cmidrule(lr){2-2} \cmidrule(lr){3-4} \cmidrule(lr){5-6} \cmidrule(lr){7-8} \cmidrule(lr){9-9}
 & FT & 75.7 & 74.3 & 75.8 & 75.9 & 73.6 & 73.8 & 74.8 \\
 & \cellcolor{gg}{Shadow-FT} & 74.9 & 75.7 & 76.1 & 76.2 & 73.5 & 73.8 & \cellcolor{gg}{\textbf{75.0}} \\
\midrule

\multirow{3}{*}{Llama-3.2-1B} & Instruct & \multicolumn{2}{c}{23.8} & \multicolumn{2}{c}{26.5} & \multicolumn{2}{c}{34.2} & 28.2 \\
\cmidrule(lr){2-2} \cmidrule(lr){3-4} \cmidrule(lr){5-6} \cmidrule(lr){7-8} \cmidrule(lr){9-9}
 & FT & 24.5 & 25.3 & 26.1 & 26.6 & 32.8 & 33.3 & 28.1 \\
 & \cellcolor{gg}{Shadow-FT} & 25.2 & 27.2 & 28.2 & 27.9 & 32.7 & 32.3 & \cellcolor{gg}{\textbf{29.0}} \\
\midrule
\multirow{3}{*}{Llama-3.2-3B} & Instruct & \multicolumn{2}{c}{53.6} & \multicolumn{2}{c}{39.3} & \multicolumn{2}{c}{49.3} & 47.4 \\
\cmidrule(lr){2-2} \cmidrule(lr){3-4} \cmidrule(lr){5-6} \cmidrule(lr){7-8} \cmidrule(lr){9-9}
 & FT & 52.7 & 51.9 & 40.2 & 41.4 & 49.4 & 49.1 & 47.5 \\
 & \cellcolor{gg}{Shadow-FT} & 54.9 & 56.2 & 40.3 & 42.8 & 49.5 & 48.9 & \cellcolor{gg}{\textbf{48.8}} \\
\midrule
\multirow{3}{*}{Llama-3.1-8B} & Instruct & \multicolumn{2}{c}{56.8} & \multicolumn{2}{c}{50.9} & \multicolumn{2}{c}{56.6} & 54.8 \\
\cmidrule(lr){2-2} \cmidrule(lr){3-4} \cmidrule(lr){5-6} \cmidrule(lr){7-8} \cmidrule(lr){9-9}
 & FT & 56.8 & 57.8 & 53.4 & 51.8 & 58.5 & 57.5 & 56.0 \\
 & \cellcolor{gg}{Shadow-FT} & 58.7 & 59.4 & 51.8 & 50.9 & 57.6 & 58.7 & \cellcolor{gg}{\textbf{56.2}} \\

\bottomrule
\end{tabular}

}
\end{table}

Table \ref{table: main_res} shows the results of tuning various mainstream LLMs on BAAI-2k using full-parameter fine-tuning and LoRA.
We set the rank as 128 in LoRA.
Some findings can be summarized as follows:
\begin{itemize}

\item \textbf{Conventional tuning methods lead to marginal improvements and even performance degeneration.}
Considering the average score, we can find that conventional tuning methods bring marginal improvements, such as 74.8 vs. 74.5 on Qwen-2.5-32B and 47.4 vs. 47.5 on Llama-3.2-3B.
Moreover, they would lead to performance degeneration, such as 68.0 vs. 65.7 on Qwen-3-4B and 70.6 vs. 69.2 on Qwen-3-8B.
The observations are consistent across full-parameter tuning and LoRA.

\item \textbf{While conventional tuning fails, Shadow-FT performs well in adaptation at the same cost.}
Across all model sizes and tasks, Shadow-FT consistently outperforms tuning baselines and the vanilla \textsc{Instruct} model.
For example, on Qwen-3-4B, Shadow-FT archives an average score of 69.4, which is 3.7 higher than the 65.7 of conventional tuning methods and 1.4 higher than the vanilla \textsc{Instruct} model.
The conclusion is consistent on larger models such as Qwen-3-14B.
Moreover, Shadow-FT does not introduce any extra training overheads.
These consistent gains demonstrate that our proposed Shadow-FT can effectively learn the knowledge contained in training data.

\item \textbf{Shadow-FT works well under both full-parameter setting and LoRA.}
For instance, when tuning Qwen-3-4B under full-parameter setting, Shadow-FT achieves 73.7/67.4/64.9 on Math-7/Code-3/Knowledge-9 compared to 72.9/66.4/62.9 of conventional tuning methods.
When applying a low-rank setting, Shadow-FT achieves 75.9/69.7/65.0, which is 4.7/10.1/3.9 higher than conventional LoRA.
These indicate that Shadow-FT is effective with different tuning strategies, showing its robustness.

\item \textbf{LoRA can outperform full-parameter.}
When tuning using our BAAI-2k dataset, we find that Shadow-FT (LoRA) can outperform Shadow-FT (full), such as 69.7 vs. 67.4 on Code-3 when tuning Qwen-3-4B.
Interestingly, we find that Shadow-FT (LoRA) typically performs better than Shadow-FT (full) on Math-7.
However, considering the conventional tuning methods, FT (full) would perform better \citep{biderman2024lora}, such as 75.9 vs. 74.8 on Qwen-3-8B.
We leave it to future work for further investigation.

\end{itemize}

Moreover, we further conduct a case study on Llama-3.1-8B-Instruct.
Please refer to Appendix \ref{app: case_study} for more details.
\section{Extensive Analysis}

\subsection{Ranks in LoRA}
\label{sec: ranks}
\begin{figure*}[t!]
\centering
\begin{minipage}[t]{0.55\textwidth}
\small
\centering
\includegraphics[width=\textwidth]{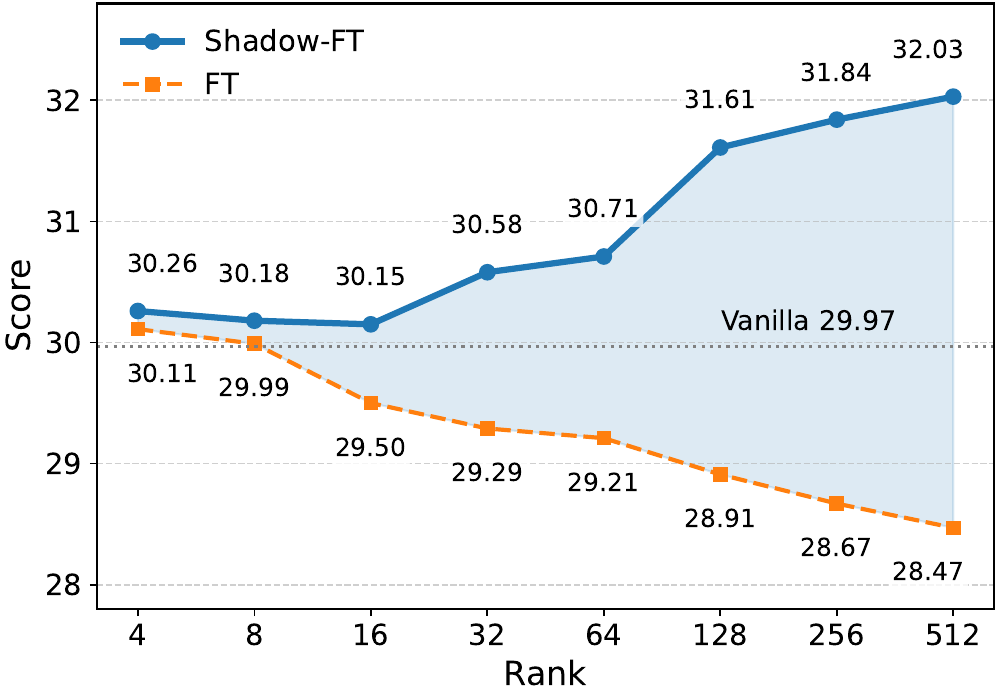}
\caption{
The average of Math-7, Code-3, and Knowledge-9 for different ranks when tuning Llama-3.2-1B using LoRA.
We report the best performance searching learning rates in \{5e-5, 1e-4, 2e-4, 5e-4\}.
}
\label{fig: rank}
\end{minipage}
\quad 
\begin{minipage}[t]{0.41\textwidth}
\centering
\includegraphics[width=\textwidth]{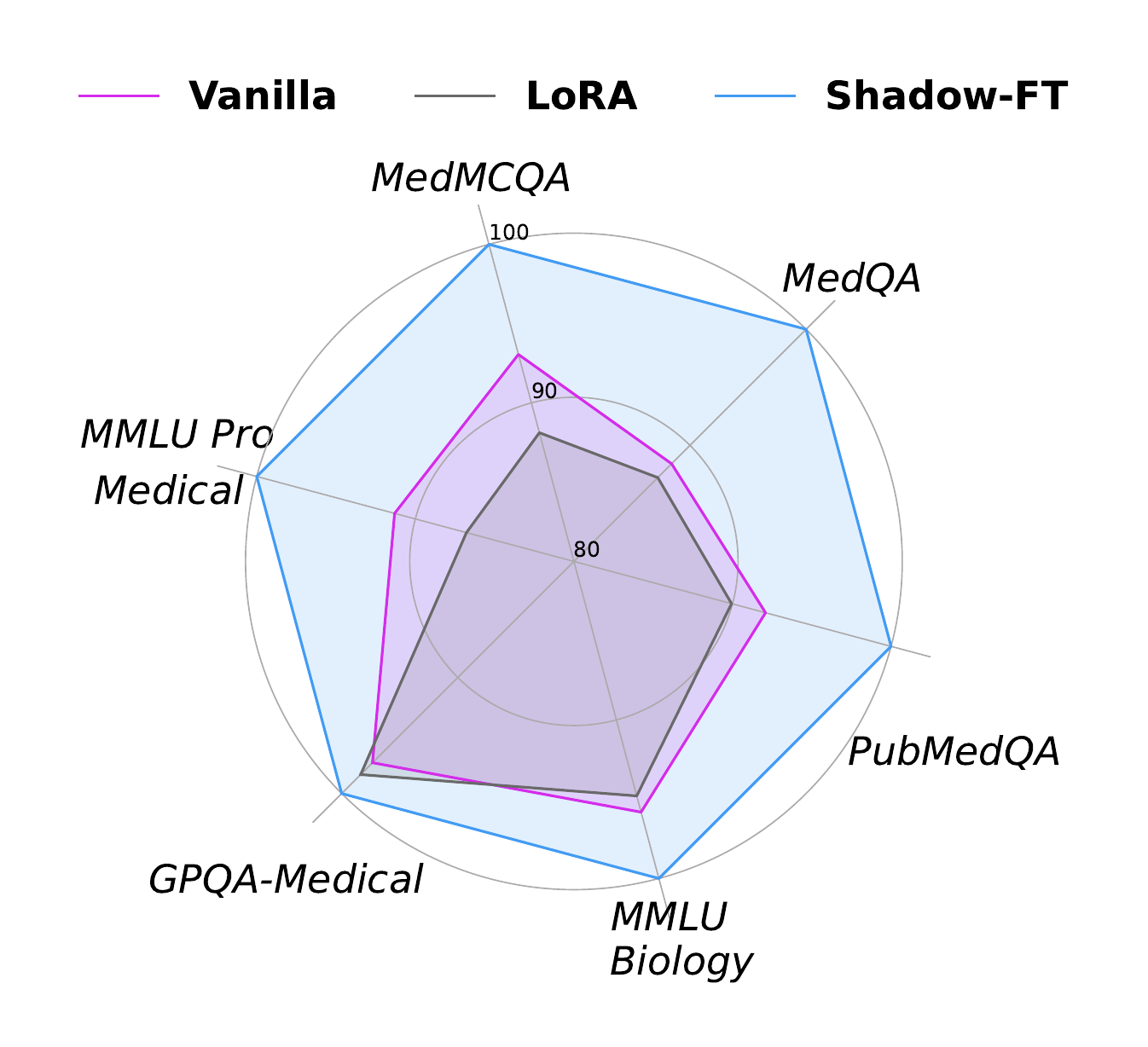}
\caption{
Performance of various methods when tuning Llama-3.2-1B on the Medical-o1-reasoning-SFT dataset. 
Detailed scores at Table \ref{table:llama3_medical}.
}
\label{fig: medical}
\end{minipage}
\end{figure*}

We fine-tune the Llama-3.2-1B using LoRA with different ranks (from 4 to 512), and report the average scores after searching learning rates in \{5e-5, 1e-4, 2e-4, 5e-4\}.
As shown in Figure \ref{fig: rank}, our proposed Shadow-FT (LoRA) can always outperform conventional LoRA with different ranks, demonstrating the robustness.
With a larger rank, the conventional LoRA would perform worse, indicating more severe degeneration when tuning the \textsc{Instruct} model~\citep{yang2024llm}.
In contrast to that, Shadow-FT (LoRA) can consistently benefit from more parameters (with larger ranks) and achieves better performance.
For the results on Llama-3.1-8B, please refer to Appendix \ref{appdx: rank}.

\subsection{Tuning on Domain Data}
Tuning methods are typically employed to adapt LLMs for a specific domain, such as medical.
Therefore, we perform tuning experiments on specific domain data, including Medical-o1-reasoning-SFT \citep{chen2024huatuogpt} in the medical domain, Code-Z1 \citep{yu2025z1} in the code domain, and LIMO \citep{ye2025limo} \& OpenR1-Math~\citep{openr1}
in the math domain.
Following LIMO~\citep{ye2025limo}, we uniformly down sample the Medical-o1-reasoning-SFT to 1,000, and Code-Z1/OpenR1-Math to 2,000.
On these domain tasks, we employ the LoRA with rank 128 and optimize with a learning rate of 2e-4.

\begin{table}[!t]
\centering
\scriptsize
\caption{The detailed mathematical and code performance tuning Qwen-3-8B and Llama-3.1-8B on Code-Z1, LIMO, and OpenR1-Math. 
\textit{Ins.} denotes the vanilla \textsc{Instruct} baseline, \textit{LoRA} for conventional LoRA, and \textit{Shadow} for proposed Shadow-FT (LoRA). 
{\color{ForestGreen}{Green$\downarrow$}}/{\color{red}Red$\uparrow$} indicates a performance drop/gain relative to the vanilla \textsc{Instruct} baseline.
}

\renewcommand{\arraystretch}{1.2}
\setlength{\tabcolsep}{1.6pt}
\begin{tabular}{l*{14}{c}}
\toprule
\multirow{3}{*}{\textbf{Benchmark}} & \multicolumn{7}{c}{\textbf{Qwen-3-8B}}
& \multicolumn{7}{c}{\textbf{Llama-3.1-8B}}\\
\cmidrule(lr){2-8}\cmidrule(lr){9-15}
& 
& \multicolumn{2}{c}{Code-Z1}
& \multicolumn{2}{c}{LIMO}
& \multicolumn{2}{c}{OpenR1-Math}
& 
& \multicolumn{2}{c}{Code-Z1}
& \multicolumn{2}{c}{LIMO}
& \multicolumn{2}{c}{OpenR1-Math}\\
\cmidrule(lr){3-4}\cmidrule(lr){5-6}\cmidrule(lr){7-8}\cmidrule(lr){10-11}
\cmidrule(lr){12-13}\cmidrule(lr){14-15}
&   \textit{Ins.}    & \textit{LoRA} & \textit{Shadow} &  \textit{LoRA} & \textit{Shadow} &  \textit{LoRA} & \textit{Shadow}
&   \textit{Ins.}    &  \textit{LoRA} & \textit{Shadow} &  \textit{LoRA} & \textit{Shadow} &  \textit{LoRA} & \textit{Shadow} \\
\cmidrule{1-1} \cmidrule{2-8} \cmidrule{9-15}

AIME24        & 20.0 & 13.3 & 36.7 & 23.3 & 26.7 & 16.7 & 26.7 &  6.7 &  3.3 & 20.0 &  6.7 &  3.3 &  3.3 &  6.7 \\
GSM8K(8shot)  & 87.4 & 84.1 & 88.3 & 85.2 & 88.7 & 83.1 & 86.8 & 84.2 & 84.1 & 85.8 & 80.5 & 83.8 & 82.3 & 84.8 \\
GSM8K(0shot)  & 93.0 & 91.9 & 93.6 & 91.7 & 92.4 & 92.7 & 92.9 & 84.2 & 85.4 & 85.7 & 82.5 & 86.1 & 86.1 & 85.9 \\
MATH          & 70.9 & 69.4 & 69.1 & 70.0 & 67.6 & 70.6 & 66.5 & 48.0 & 48.8 & 51.3 & 44.3 & 45.8 & 39.8 & 47.7 \\
MATH-500      & 83.2 & 79.8 & 88.0 & 77.0 & 80.4 & 80.2 & 85.0 & 48.4 & 50.8 & 55.4 & 44.4 & 43.8 & 41.8 & 48.8 \\
Minerva\_Math & 73.0 & 69.7 & 72.9 & 69.9 & 73.1 & 70.8 & 73.2 & 40.6 & 39.6 & 45.5 & 37.1 & 41.2 & 44.0 & 44.2 \\
SVAMP         & 91.4 & 90.3 & 93.3 & 90.9 & 92.9 & 90.3 & 93.0 & 83.1 & 86.5 & 86.9 & 83.7 & 85.9 & 85.1 & 87.1 \\

\cmidrule(lr){2-2}\cmidrule(lr){3-4}\cmidrule(lr){5-6}\cmidrule(lr){7-8}\cmidrule(lr){9-9}\cmidrule(lr){10-11}
\cmidrule(lr){12-13}\cmidrule(lr){14-15}

\textbf{Math-7} 
& \gut{gray!20}{74.5\(\,\diamond\)}
& \gut{white}{71.2\(\,\textcolor{ForestGreen}{\downarrow}\)} 
& \gut{red!15}{\textbf{77.4}\(\,\textcolor{red}{\uparrow}\)} 
& \gut{white}{72.6\(\,\textcolor{ForestGreen}{\downarrow}\)} 
& \gut{red!15}{\textbf{75.1}\(\,\textcolor{red}{\uparrow}\)} 
& \gut{white}{72.1\(\,\textcolor{ForestGreen}{\downarrow}\)} 
& \gut{red!15}{\textbf{75.7}\(\,\textcolor{red}{\uparrow}\)} 
& \gut{gray!20}{56.8\(\,\diamond\)}
& \gut{red!15}{57.1\(\,\textcolor{red}{\uparrow}\)} 
& \gut{red!15}{\textbf{61.5}\(\,\textcolor{red}{\uparrow}\)} 
& \gut{white}{54.2\(\,\textcolor{ForestGreen}{\downarrow}\)} 
& \gut{white}{\textbf{55.7}\(\,\textcolor{ForestGreen}{\downarrow}\)} 
& \gut{white}{54.6\(\,\textcolor{ForestGreen}{\downarrow}\)} 
& \gut{red!15}{\textbf{58.0}\(\,\textcolor{red}{\uparrow}\)} 
\\

\cmidrule{1-1} \cmidrule{2-8} \cmidrule{9-15}

HumanEval     & 84.2 & 82.3 & 87.8 & 84.2 & 86.0 & 78.1 & 83.5 & 71.3 & 64.6 & 70.1 & 68.9 & 70.7 & 72.6 & 72.0 \\
HumanEval+    & 79.9 & 76.8 & 78.1 & 79.3 & 81.1 & 75.6 & 81.1 & 63.4 & 48.2 & 64.6 & 62.2 & 64.0 & 61.6 & 62.8 \\
{\tiny LiveCodeBench} & 51.5 & 43.2 & 54.7 & 48.7 & 53.1 & 47.6 & 54.6 & 19.8 & 11.8 & 20.5 & 18.6 & 20.7 & 15.6 & 19.9 \\

\cmidrule(lr){2-2}\cmidrule(lr){3-4}\cmidrule(lr){5-6}\cmidrule(lr){7-8}\cmidrule(lr){9-9}\cmidrule(lr){10-11}
\cmidrule(lr){12-13}\cmidrule(lr){14-15}

\textbf{Code-3}
& \gut{gray!20}{72.7\(\,\diamond\)}  
& \gut{white}{67.4\(\,\textcolor{ForestGreen}{\downarrow}\)}
& \gut{red!15}{\textbf{73.5}\(\,\textcolor{red}{\uparrow}\)}
& \gut{white}{70.7\(\,\textcolor{ForestGreen}{\downarrow}\)}
& \gut{red!15}{\textbf{73.4}\(\,\textcolor{red}{\uparrow}\)}
& \gut{white}{67.1\(\,\textcolor{ForestGreen}{\downarrow}\)}
& \gut{red!15}{\textbf{73.1}\(\,\textcolor{red}{\uparrow}\)}
& \gut{gray!20}{50.9\(\,\diamond\)}  
& \gut{white}{41.5\(\,\textcolor{ForestGreen}{\downarrow}\)}
& \gut{red!15}{\textbf{51.7}\(\,\textcolor{red}{\uparrow}\)}
& \gut{white}{49.9\(\,\textcolor{ForestGreen}{\downarrow}\)}
& \gut{red!15}{\textbf{51.8}\(\,\textcolor{red}{\uparrow}\)}
& \gut{white}{49.9\(\,\textcolor{ForestGreen}{\downarrow}\)}
& \gut{red!15}{\textbf{51.6}\(\,\textcolor{red}{\uparrow}\)}
\\

\bottomrule
\end{tabular}
\label{tab:domain}
\end{table}

Figure \ref{fig: medical} reports the results of tuning Llama-3.2-1B on Medical-o1-reasoning-SFT.
We report the results on MMLU Pro-Medical \citep{wang2024mmlu}, MedMCQA \citep{pal2022medmcqa}, PubMedQA \citep{jin2019pubmedqa}, MMLU-Biology \citep{yue2024mmmu}, and GPQA-Medical following \citep{chen2024huatuogpt}, while normalizing the maximum score to 1 for better visualization.
We can find that conventional LoRA would lead to performance degeneration, while Shadow-FT (LoRA) improves the performance, which is consistent with the conclusion on BAAI-2k.
Besides, we further report the results directly tuning the \textsc{Base}.
Please refer to Appendix \ref{appdix: medical_details} for detailed scores.

Table~\ref{tab:domain} shows the detailed results of Math-7 and Code-3 tuning Qwen-3-8B and Llama-3.1-8B on Code-Z1, LIMO, and OpenR1-Math.
The observations are consistent, i.e., conventional LoRA would lead to degeneration, while the proposed shadow-FT (LoRA) can effectively adapt LLMs on specific domain knowledge.
For instance, Shadow-FT (LoRA) achieves a Math-7 score of 77.4 on Qwen-3-8B, which is 6.2 higher than 71.2 of LoRA and 2.9 higher than the vanilla \textsc{Instruct} model.
Moreover, we also find that tuning LLM via Shadow-FT on code data can improve the math capability \citep{yu2025z1}, and vice versa.
In particular, when tuned via shadow-FT on Code-z1, the Qwen-3-8B can achieve a score of 36.7 on the tough AIME-24 benchmark, showing superior adaptation and generalization ability.

\subsection{Mechanistic Analysis of Optimization Dynamics}

\begin{figure*}[t!]
\centering
\includegraphics[width=\textwidth]{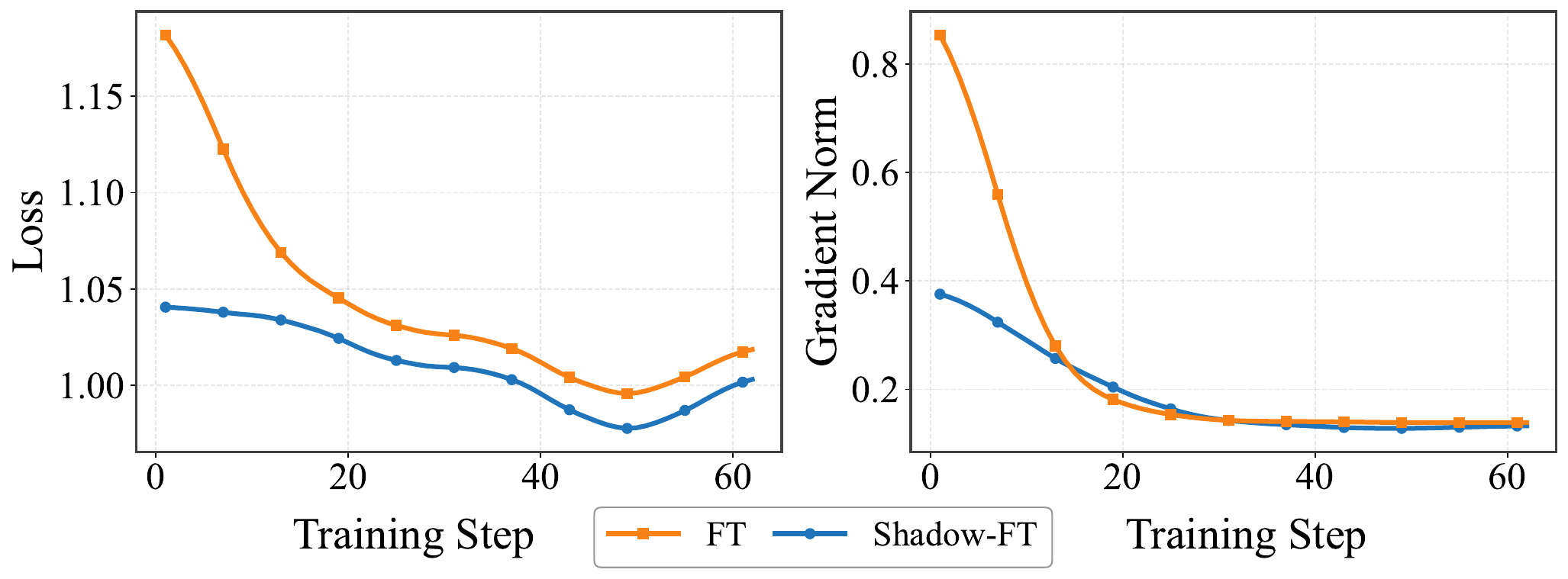}
\caption{
Optimization dynamics on loss and gradient when tuning Qwen3-8B via \textsc{Instruct}~(i.e., FT) and \textsc{Base}~(i.e., Shadow-FT).
}
\label{fig:optimization_dynamics}
\end{figure*}

To provide insight into why Shadow-FT outperforms vanilla FT, we further analyze the optimization dynamics of both methods from a loss and gradient perspective. 
We denote the loss and gradient for the \textsc{INSTRUCT} model (tuned with vanilla FT) as \textbf{Loss(I)} and \textbf{Grad(I)}, and for the BASE model (tuned with Shadow-FT) as \textbf{Loss(S)} and \textbf{Grad(S)}.

Figure~\ref{fig:optimization_dynamics} illustrates these metrics while tuning Qwen3-8B with LoRA (r=128). The two approaches show markedly different training dynamics.
\textit{At initialization}, \textbf{Loss(I)} is 22.6\% higher than \textbf{Loss(S)}, and \textbf{Grad(I)} is 3.25$\times$ larger, reflecting a poor task fit and strong resistance from the instruction-following prior~\citep{ji2024language}. 
\textit{During training}, \textbf{Grad(I)} decays precipitously (11/61 step), while \textbf{Grad(S)} decreases more moderately (58\%), indicating that \textsc{Instruct} quickly enters a rigid optimization regime with suppressed updates while Shadow-FT sustains enjoying a smoothed learning. 
\textit{By convergence} (step 61/61), both gradients stabilize at similar magnitudes, but \textbf{Loss(I)} remains 2.4\% higher, evidencing inferior adaptation.

Overall, the figure reveals a fundamental contrast in optimization.
Vanilla FT on the \textsc{Instruct} model is hindered by a large initial gradient that rapidly diminishes, while the \textsc{Base} model can avoid this and shows a stable trajectory.
However, \textsc{Base} model is a good learner but a poor backbone due to the lack of post-training.
The conclusion is consistent with Pass@k results detailed in Appendix~\ref{appendix: pass@k}.
Therefore, we propose Shadow-FT to learn on \textsc{Base} and execute on \textsc{Instruct}.

\subsection{Combined with DPO}
Direct Preference Optimization~(DPO), which directly optimizes a language model to adhere to human preferences without explicit reward modeling or reinforcement learning, shows promising performance when applying RL to LLMs \citep{rafailov2023direct}.
Therefore, we try to combine Shadow-FT with DPO, i.e., applying DPO on \textsc{Base} and then grafting the weight to \textsc{Instruct}, termed as Shadow-DPO.
Specifically, we achieve Shadow-DPO using LoRA on 1,000 paired samples from the Math-Step \citep{lai2024step} dataset and set the rank to 8 and 128.
As shown in Table \ref{res: dpo}, shadow-DPO outperforms DPO under two settings, such as 55.39 vs. 54.62 of vanilla DPO.
It shows that the strategy employing the \textsc{Base} as proxy of \textsc{Instruct} also works for DPO.
Meanwhile, a larger rank leads to better results for shadow-DPO, which is consistent with results tuning on BAAI-2k shown in Figure \ref{fig: rank}.





\subsection{Performance on MLLM}

For generality, we further conduct experiments tuning Multimodal Large Language Models (MLLMs).
For the dataset, we select 10,000 samples from ChartMoE ~\citep{ChartMoE}, which takes a chart and a natural language question as input to predict the answer.
For MLLM, we select Gemma-3 \citep{team2025gemma} 12B/27B and Llama-3.2-Vision \citep{grattafiori2024llama} 11B/90B.
During training, we employ LoRA and set the rank to 128.
The learning rate is 2e-4.
We evaluate the tuned model via lmms-eval framework \citep{zhang2024lmmsevalrealitycheckevaluation}.
As shown in Table \ref{table:MLLM}, both conventional LoRA and Shadow-FT (LoRA) effectively adapt MLLMs on ChartQA~\citep{masry2022chartqa} tasks.
Meanwhile, our proposed Shadow-FT outperforms LoRA, especially on larger models, such as 63.80 on Gemma-3-27B compared to 60.28 of vanilla LoRA and 80.6 on Llama-3.2-Vision-90B compared to 79.92.

\begin{figure*}[t!]
\centering
\begin{minipage}[t]{0.40\textwidth}
\small
\centering
\captionof{table}{
Performance of tuning Llama-3.1-8B using DPO and Shadow-DPO with ranks be 8 and 128, respectively.
}
\label{tab:metrics}
\begin{tabular}{lcc}
\toprule
\textbf{Method} & \textbf{Rank} & \textbf{19-Avg.} \\
\midrule
Vanilla & --  & 54.77 \\
\midrule
DPO     & 8   & 54.80 \\
Shadow-DPO & 8   & \textbf{54.96}  \\
\midrule
DPO     & 128 & 54.62 \\
Shadow-DPO & 128 & \textbf{55.39} \\
\bottomrule
\end{tabular}
\label{res: dpo}

\end{minipage}
\quad 
\begin{minipage}[t]{0.56\textwidth}
\centering
\captionof{table}{
Performance of Gemma-3 and Llama-3.2-Vision on the multi-modal ChartQA task.
We set the rank of LoRA to 128.
}
\renewcommand{\arraystretch}{1.2}
\begin{tabular}{llccc}
\toprule
\textbf{Model} & \textbf{Size} & \textbf{Vanilla} & \textbf{LoRA} & \textbf{Shadow-FT} \\
\midrule
\multirow{2}{*}{Gemma-3}
  & 12B & 37.36 & 53.48 & \textbf{54.92} \\
  & 27B & 41.92 & 60.28 & \textbf{63.80} \\
\midrule
Llama-3.2
  & 11B & 22.12 & \textbf{74.44} & 74.12 \\
-Vision & 90B & 30.92 & 79.92 & \textbf{80.60} \\
\bottomrule
\end{tabular}
\label{table:MLLM}

\end{minipage}
\end{figure*}

\subsection{Weight Delta Scaling}
In Shadow-FT, we directly graft the learned weights from \textsc{BASE} to \textsc{Instruct}.
We further explore the scaling of learned weights.
Please refer to Appendix \ref{appendix: weight_delta} for more details.
In summary, our proposed Shadow-FT outperforms vanilla \textsc{Instruct} with different scaling factors, showing strong robustness.
Moreover, a factor slightly larger than 1 would yield better results, while we leave it for future work to explore the best factor.

\subsection{Model Zoo: More LLMs}
\label{sec: zoo}

We further apply Shadow-FT to more LLMs, including Gemma-3 series \citep{team2025gemma}, Yi series \citep{young2024yi}, and Falcon series \citep{almazrouei2023falcon}.
Please refer to Appendix \ref{apx: zoo} for more details.
We can find that proposed Shadow-FT consistently outperforms conventional tuning methods.
All the tuned models will be made public in the future.

\section{Related Work}

\subsection{Tuning For LLMs}
Large language models (LLMs) gain superior ability from pre-training on tremendous data~\citep{gururangan2020don}, followed by tuning on various downstream tasks~\citep{ouyang2022training, muennighoff2025s1}.
These methods can be categorized into: 1) full-parameters method, which updates all the parameters, and 2) parameter-efficient fine-tuning~(PEFT) method, lowering the tuning costs via parameter selection~\citep{zaken2021bitfit} or low-rank branches~\citep{hu2022lora, wu2024mixture}.
More recently, Reinforcement Learning from Human Feedback~(RLHF) methods show promising performance in aligning models to human preferences and improving the reasoning ability \cite{rafailov2023direct, bai2023qwen, guo2025deepseek, team2025kimi}.
These methods focus on improving the training strategy and involve the target model only.
In this paper, we propose Shadow-FT to tune \textsc{Instruct} model on \textsc{Base} model. 
Also, our proposed Shadow-FT can be combined with these baselines to enhance the performance.

\subsection{Model Guidance in Tuning}

Introducing extra knowledge from other models has been proven as a promising way to enhance tuning performance, such as knowledge distillation~\citep{hinton2015distilling, wu2024rethinking} and proxy-tuning~\citep{liu2024tuning}.
Knowledge distillation methods aim to transfer the knowledge from a larger teacher model to a compact student model, via aligning the outputs~\citep{wu2024rethinking, yang2024loca} or employing the teacher's outputs as training data~\citep{qin2024o1, min2024imitate}.
Proxy-tuning first tunes a smaller LLM and then applies the logit differences to a larger model~\citep{liu2024tuning}.
These methods transfer knowledge at the feature level or data level, while our proposed Shadow-FT directly grafts the weight updates.
RE-Adapt~\citep{DBLP:journals/corr/abs-2405-15007} also utilizes the Base/Instruct model pair for adaptation.
However, RE-Adapt models the static weight difference with a \textit{low-rank} approximation, whereas Shadow-FT is a model-free approach that directly transfers the full dynamic updates without any assumption.
Additionally, we notice a very recent concurrent work \citep{lin2025efficient} to transfer the fine-tuning ability.
Differently, our proposed Shadow-FT focuses on tuning \textsc{Instruct} via \textsc{Base} model based on the \textit{observation} that the weights are highly similar.
Moreover, we conduct experiments on more LLMs across more benchmarks, and further extend the idea to MLLMs and DPO.

\section{Conclusion}

In this work, we propose Shadow-FT, a novel framework to fine-tune \textsc{Instruct} models by leveraging their corresponding \textsc{Base} models. 
Inspired by the observation that the weights of \textsc{Base} and \textsc{Instruct} are highly similar, we propose Shadow-FT to tune \textsc{Instruct} vis \textsc{Base}, aiming to tune \textsc{Instruct} better.
Extensive experiments across multiple LLM series, including Qwen, Llama, Gemma, and Falcon, demonstrate that Shadow-FT consistently outperforms conventional full-parameter and parameter-efficient fine-tuning methods.
Notably, Shadow-FT introduces no additional training cost or parameters, yet it achieves superior performance across diverse benchmarks covering math, coding, and reasoning tasks. 
We further show that Shadow-FT generalizes well to multimodal large language models (MLLMs) and can be seamlessly combined with alignment techniques such as DPO, offering a simple yet effective solution for improving instruction-following models.

\section*{Reproducibility Statement}
We are committed to the reproducibility of our work. 
The MATH-7 results, averaged across three trials, are shown in Table~\ref{table: main_res} and Table~\ref{tab:scaling}.
The full source code required to reproduce our experiments is included in the supplementary material. 
Corresponding hyperparameters and detailed configuration files for all experiments are documented in Section~\ref{Exp:setting}. 
All experiments were conducted on publicly available benchmarks, and the details are provided in Appendix~\ref{appendix: benchmark}.

\section*{Ethics statement}

We have adhered to the ICLR Code of Ethics in this research.
Our work is based entirely on publicly available models and benchmarks, involves no human subjects, and we commit to releasing our code to ensure reproducibility.


\appendix
\newpage

\appendix

\section*{LLM Usage Disclosure}
The human authors are primarily responsible for this work. We utilized several large language models (e.g., GPT-4, Gemini Pro, Claude 3) as general-purpose assistive tools to improve the quality of our research and writing. Their use was limited to the following specific tasks: assisting with code implementation and debugging, generating boilerplate code, refining the language and formatting of the manuscript, and proofreading. The authors conceived all research ideas, designed the experiments, and performed the final analysis of the results. We take full responsibility for all content in this paper and confirm that it complies with relevant licenses and ethical guidelines.

\section{Benchmarks Details}
\label{appendix: benchmark}

\begin{table*}[!h]
\centering
\caption{Details on instruction-model evaluations. 
CoT denotes the chain-of-thought setting.
}
\vspace{1em}

\begin{tabular}{l c c c c}
\toprule
\textbf{Evaluation}        & \textbf{Metric} & \textbf{Type} & \textbf{n-shot}  & \textbf{CoT}\\ 

\midrule
\multicolumn{5}{c}{\textit{Math-7}}\\
\midrule

\textbf{AIME24}            &   pass@1                  & sampling  &  0-shot  &       \\           
\textbf{GSM8K(0-shot)}          &   Accuracy                & sampling  &  0-shot  &  \cmark  \\ 
\textbf{GSM8K(8-shot)}          &   Accuracy                & sampling  &  8-shot  &  \cmark  \\   

\textbf{MATH}           &   Accuracy                & sampling  &  0-shot  &      \cmark \\          \textbf{MATH-500}           &   Accuracy                & sampling  &  0-shot  &       \\           

\textbf{Minerva Math}     &   Accuracy                & sampling  &  4-shot  &       \\           
\textbf{SVAMP}          &   Accuracy                & sampling  &  0-shot  &    \\

\midrule
\multicolumn{5}{c}{\textit{Code-3}}\\
\midrule

\textbf{HumanEval}      &   pass@1                  & sampling  &  0-shot  &       \\           
\textbf{HumanEval+}      &   pass@1                  & sampling  &  0-shot  &       \\     
\textbf{LiveCodeBench}  &   &  average &    &    \\              
{       - generation}  &   pass@1  & sampling  &  0-shot  &  \cmark  \\              
{  - test}  &   pass@1  & sampling  &  0-shot  &  \cmark  \\              
{   - prediction}  &   pass@1  & sampling  &  0-shot  &  \cmark  \\              

\midrule
\multicolumn{5}{c}{\textit{Knowledge-9}}\\
\midrule
\textbf{ARC-Challenge}            &   Accuracy                & sampling  &  0-shot  &   \cmark    \\        

\textbf{BBH(0-shot)}            &   Accuracy                & sampling  &  0-shot  &       \\          
\textbf{BBH(3-shot)}            &   Accuracy                & sampling  &  3-shot  &       \\         
\textbf{Drop}            &   Accuracy                & sampling  &  0-shot  &       \\        

\textbf{GPQA Diamond}    & Accuracy                & sampling   & 0-shot    & \cmark    \\ 

\textbf{MMLU}           &   Accuracy                & sampling  &  0-shot  &       \\           
\textbf{MMLU Pro}           &   Accuracy                & sampling  &  0-shot  &       \\           

\textbf{Winogrande}            &   Accuracy                & sampling  &  0-shot  &       \\          
\textbf{TheoremQA}            &   Accuracy                & sampling  &  0-shot  &       \\

\bottomrule
\end{tabular}

\label{tab:eval_detail_it}
\end{table*}

The details about the benchmarks are detailed in Table \ref{tab:eval_detail_it}.
Since the n-shot setting are unstable, we prefer to report the 0-shot results.
For the popular GSM8K and BBH, we also report 8-shot and 3-shot results.

\section{Model Zoo: More LLMs}
\label{apx: zoo}

\begin{table}[t]
\centering
\caption{Performance comparison of different methods tuning more LLMs. 
We employ the Instruct version and report the final average scores.
}
\renewcommand{\arraystretch}{1.1}
\begin{tabular}{ccccccccc}
\toprule
\label{table: model_zoo}

\multirow{2}{*}{\textbf{Model}} & \multirow{2}{*}{\textbf{Method}} & \multicolumn{2}{c}{\textbf{Math-7}} & \multicolumn{2}{c}{\textbf{Code-3}} & \multicolumn{2}{c}{\textbf{Knowledge-9}} & \multirow{2}{*}{\textbf{Avg.}} \\
\cmidrule(lr){3-4}\cmidrule(lr){5-6}\cmidrule(lr){7-8}
 &  & \textit{Full} & \textit{LoRA} & \textit{Full} & \textit{LoRA} & \textit{Full} & \textit{LoRA} \\
\midrule
\addlinespace
 &  \multicolumn{6}{c}{\textit{Falcon Family}}\\
\addlinespace
\midrule
\multirow{3}{*}{Falcon3-3B} & Vanilla & \multicolumn{2}{c}{53.33} & \multicolumn{2}{c}{38.09} & \multicolumn{2}{c}{47.28} & 46.23 \\
\cmidrule(lr){2-2}\cmidrule(lr){3-4}\cmidrule(lr){5-6}\cmidrule(lr){7-8}\cmidrule(lr){9-9}
 & FT & 55.83 & 58.70 & 39.57 & 41.23 & 48.43 & 49.50 & 48.88 \\
 & \cellcolor{gg}{Shadow-FT} & 56.74 & 60.31 & 41.02 & 43.69 & 48.16 & 48.25 & \cellcolor{gg}{\textbf{49.70}} \\
\midrule
\multirow{3}{*}{Falcon3-10B} & Vanilla & \multicolumn{2}{c}{57.23} & \multicolumn{2}{c}{60.03} & \multicolumn{2}{c}{53.85} & 57.04 \\
\cmidrule(lr){2-2}\cmidrule(lr){3-4}\cmidrule(lr){5-6}\cmidrule(lr){7-8}\cmidrule(lr){9-9}
 & FT & 59.33 & 68.74 & 60.95 & 61.54 & 54.17 & 55.72 & \textbf{60.08} \\
 & \cellcolor{gg}{Shadow-FT} & 58.27 & 70.40 & 61.35 & 62.20 & 53.19 & 52.83 & \cellcolor{gg}{59.71} \\

\midrule
\addlinespace
 & \multicolumn{6}{c}{\textit{Gemma Family}}\\
\addlinespace
\midrule
\multirow{3}{*}{Gemma-3-4B} & Vanilla & \multicolumn{2}{c}{54.02} & \multicolumn{2}{c}{48.33} & \multicolumn{2}{c}{52.01} & 51.45 \\
\cmidrule(lr){2-2}\cmidrule(lr){3-4}\cmidrule(lr){5-6}\cmidrule(lr){7-8}\cmidrule(lr){9-9}
 & FT & 35.34 & 49.12 & 48.15 & 43.03 & 43.83 & 48.37 & 44.64 \\
 & \cellcolor{gg}{Shadow-FT} & 56.68 & 56.30 & 48.87 & 48.93 & 52.88 & 51.62 & \cellcolor{gg}{\textbf{52.55}} \\
\midrule

\multirow{3}{*}{Gemma-3-12B} & Vanilla & \multicolumn{2}{c}{60.82} & \multicolumn{2}{c}{58.06} & \multicolumn{2}{c}{61.54} & 60.14 \\
\cmidrule(lr){2-2}\cmidrule(lr){3-4}\cmidrule(lr){5-6}\cmidrule(lr){7-8}\cmidrule(lr){9-9}
 & FT & 56.56 & 62.84 & 58.17 & 59.21 & 61.63 & 61.99 & 60.07 \\
 & \cellcolor{gg}{Shadow-FT} & 61.05 & 64.59 & 58.17 & 60.86 & 61.59 & 62.66 & \cellcolor{gg}{\textbf{61.49}} \\
\midrule
\addlinespace
 &  \multicolumn{6}{c}{\textit{Yi Family}}\\
\addlinespace
\midrule
\multirow{3}{*}{Yi-6B} & Vanilla & \multicolumn{2}{c}{17.34} & \multicolumn{2}{c}{8.40} & \multicolumn{2}{c}{38.63} & 21.46 \\
\cmidrule(lr){2-2}\cmidrule(lr){3-4}\cmidrule(lr){5-6}\cmidrule(lr){7-8}\cmidrule(lr){9-9}
 & FT & 18.93 & 18.39 & 10.64 & 11.89 & 40.84 & 40.46 & \textbf{23.53} \\
 & \cellcolor{gg}{Shadow-FT} & 17.73 & 17.21 & 13.35 & 14.30 & 38.70 & 38.25 & \cellcolor{gg}{23.26} \\
\midrule
\multirow{3}{*}{Yi-Coder-9B} & Vanilla & \multicolumn{2}{c}{28.01} & \multicolumn{2}{c}{61.85} & \multicolumn{2}{c}{40.73} & 43.53 \\
\cmidrule(lr){2-2}\cmidrule(lr){3-4}\cmidrule(lr){5-6}\cmidrule(lr){7-8}\cmidrule(lr){9-9}
 & FT & 26.05 & 26.11 & 52.70 & 53.95 & 39.55 & 37.22 & 39.26 \\
 & \cellcolor{gg}{Shadow-FT} & 28.41 & 29.09 & 62.07 & 64.72 & 40.27 & 39.88 & \cellcolor{gg}{\textbf{44.07}} \\
\bottomrule
\end{tabular}
\end{table}

We further apply Shadow-FT to more LLMs, including Gemma-3 series \citep{team2025gemma}, Yi series \citep{young2024yi}, and Falcon series \citep{almazrouei2023falcon}.
The hyperparameters are the same as tuning Qwen 3 and Llama 3.
Table \ref{table: model_zoo} shows the results of Math-7, Code-3, and Knowledge-9.
We can find that proposed Shadow-FT consistently outperforms conventional tuning methods.
For instance, Shadow-FT gets an average of 52.55 when tuning Gemma-3-4B, which is 1.1 higher than the vanilla \textsc{Instruct} model and 7.91 higher than conventional tuning methods.

\section{Similarity on More LLMs}
\label{appendix: more_simi}

\begin{figure*}[!h]
\centering
\includegraphics[width=\textwidth]{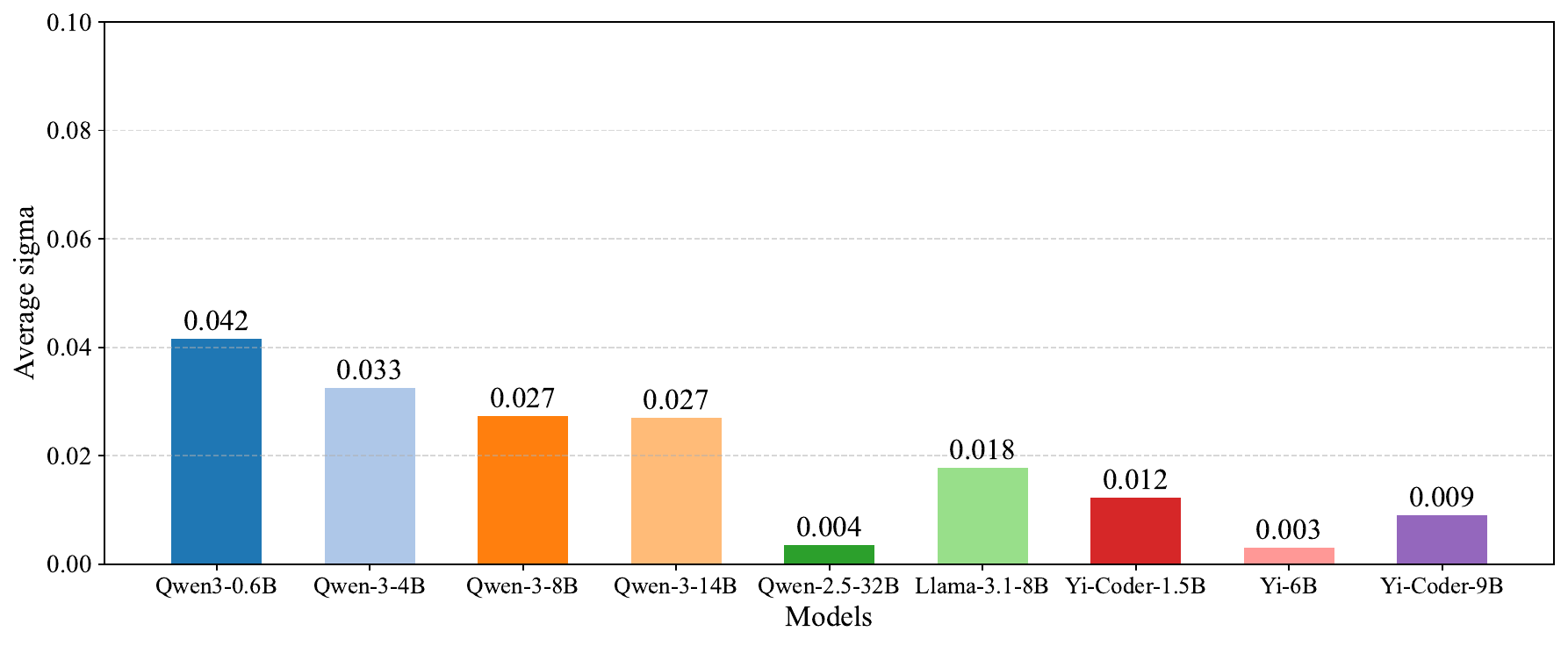}
\caption{
Average $\sigma$ values of more LLMs.
}
\label{fig: sigma_dis}
\end{figure*}

As shown in Figure \ref{fig: sigma_dis}, we can find that all $\sigma<0.05$, indicating high similarity between \textsc{Base} and \textsc{Instruct}.
Also, the larger the LLMs, the smaller the gaps.

\section{Experimental Details}

\subsection{Hyper-parameters}
\label{appendix: hyper}

For the experiments, we set the hyperparameters after grid search.
The batch size is 2, with the gradient\_accumulation\_steps as 16.
During experiments, the cutoff the inputs to 4096 and train for 1 epoch.

\subsection{Detailed Data of Table 1}

\label{appendix: detail_data}

The detailed scores are listed in Table \ref{appe_math}, Table \ref{tab:appe_code}, and Table \ref{appendix_reasoning}.

\subsection{Detailed table for Medical Benchmarks}
\label{appdix: medical_details}

Table \ref{table:llama3_medical} reports the detailed results of tuning Llama-3.2-1B on the Medical-o1-reasoning-SFT dataset.
Shadow-FTR donates the method integrating fine-tuned weights from INSTRUCT to BASE.

\begin{table}[!h]
\centering
\caption{Performance of \textsc{Llama-3.2-1B-Instruct} on medical QA benchmarks.}
\vspace{1em}

\renewcommand{\arraystretch}{1.15}
\setlength{\tabcolsep}{5pt}
\begin{tabular}{lccS@{\hspace{\seriesgap}}BBB}
\toprule
\multirow{2}{*}{\textbf{Benchmark}} & \multicolumn{3}{c}{\textbf{Tune on Instruct}} & \multicolumn{3}{c}{\textbf{Tune on Base}} \\
\cmidrule(l{3pt}r{12pt}){2-4}\cmidrule(l{-5pt}r{0pt}){5-7}
  & \cellcolor{gglight}\textbf{Instruct}
  & \cellcolor{gglight}\textbf{FT}
  & \cellcolor{ggdark}\textbf{Shadow-FT}
  & \cellcolor{ggtwo}\textbf{Base}
  & \cellcolor{ggtwo}\textbf{Base-FT}
  & \cellcolor{ggtwo}\textbf{Shadow-FTR} \\
\midrule
GPQA-Medical        & 23.85 & 24.10 & 24.50 & \textbf{25.25} & 25.00 & 24.75 \\
MMLU Pro-Medical    & 25.20 & 23.95 & \textbf{27.60} & 13.10 & 12.60 & 12.30 \\
MedMCQA             & 30.15 & 28.55 & \textbf{32.40} & 30.20 & 30.60 & 29.50 \\
MedQA               & 25.95 & 25.60 & 29.35 & 29.80 & \textbf{30.35} & 29.45 \\
PubMedQA            & 55.85 & 54.55 & \textbf{60.65} & 49.35 & 51.90 & 50.20 \\
\midrule
Avg.                & 32.20 & 31.35 & \textbf{34.90} & 29.54 & 30.09 & 29.24 \\
\bottomrule
\end{tabular}
\label{table:llama3_medical}
\end{table}

\subsection{Performance on Pass@k}
\label{appendix: pass@k}

To evaluate the exploration capability, we use the popular Pass@k, which is defined as the fraction of problems for which at least one correct response is produced in $k$ independent trials.
However, directly computing Pass@k using only $k$ rollouts for each problem often suffers from high variance.
Therefore, we adapt the unbiased estimator~\citep{chen2021evaluating}.
Specifically, we roll out for $n$ times~($n \ge k$), and calculate Pass@k as follows:
\begin{equation}
\text{Pass}@k := \mathbb{E}_{x \sim \mathcal{D}} \left[ 1 - \frac{\binom{n-c}{k}}{\binom{n}{k}} \right],
\end{equation}
where $x$ is the input prompt from dataset $D$, and $c$ is the count of correct solutions.

\begin{figure*}[!h]
\centering
\includegraphics[width=1\textwidth]{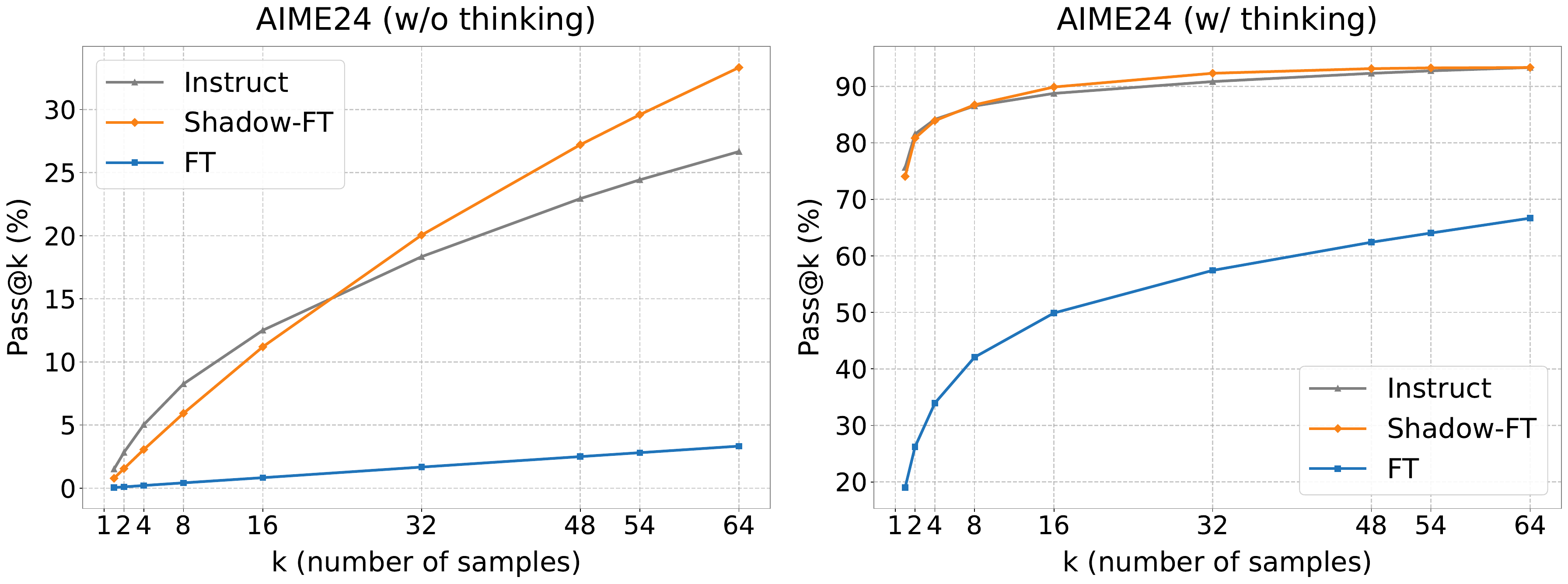}
\caption{
Pass@k performance on Qwen3-8B thinking and non-thinking modes.
}
\label{fig: pass@k}
\end{figure*}
We employ the \textsc{Qwen3-8B} model, which supports seamless switching between \emph{Thinking} mode and \emph{Non-thinking} mode.
We can easily switch between two modes using one control hyperparameter.
We set other hyperparameters following the official report of Qwen3~\citep{qwen3}:
\text{do\_sample=True}, \text{temperature=0.6}, \text{top\_k=20}, \text{top\_p=0.95}, \text{max\_new\_tokens=38912} for bettr alignment.

\paragraph{Non-thinking mode.}
Under non-thinking decoding, absolute \textsc{Pass@k} values are small for all methods, yet \textsc{Shadow-FT} exhibits a clearer upward trend with larger $k$, progressively surpassing the \textsc{Instruct} baseline. 
In contrast, vanilla \textsc{FT} yields weak performance and rarely produces correct solutions. 
Importantly, all methods used the same number of training examples and the same training cutoff length; the only difference is the initialization (\textsc{Base} vs.\ \textsc{Instruct}). 
This comparison suggests that \textsc{Base} is a better learner for supervised adaptation—its newly acquired knowledge is less prone to collapse.

\paragraph{Thinking mode.}
Under thinking mode, the effect is more pronounced: \textsc{Shadow-FT} and \textsc{Instruct} follow similarly steep, rapidly saturating curves that nearly reach the model’s capacity limit, whereas vanilla \textsc{FT} maintains a large ($>$30\%) gap across $k$. 
This pattern implies that vanilla \textsc{FT} hurts the thinking ability of \textsc{Instruct} and makes it less receptive to new knowledge compared with \textsc{Base}-initialized training.

Across both modes, our proposed Shadow-FT avoids collapse and retains—often enhances the upper-bound competence of the underlying \textsc{Instruct} model. 
This property is valuable for subsequent RL or other generalization-critical settings~\citep{zhu2025surprising}.
We attribute the robustness to the favorable inductive characteristics of \textsc{Base}-initialized learning, whereas vanilla \textsc{FT} on an \textsc{Instruct} model struggles to achieve the same balance of stability and adaptability.

\subsection{Ranks in LoRA on Llama-3.1-8B}
\label{appdx: rank}

\begin{figure*}[!h]
\centering
\includegraphics[width=0.75\textwidth]{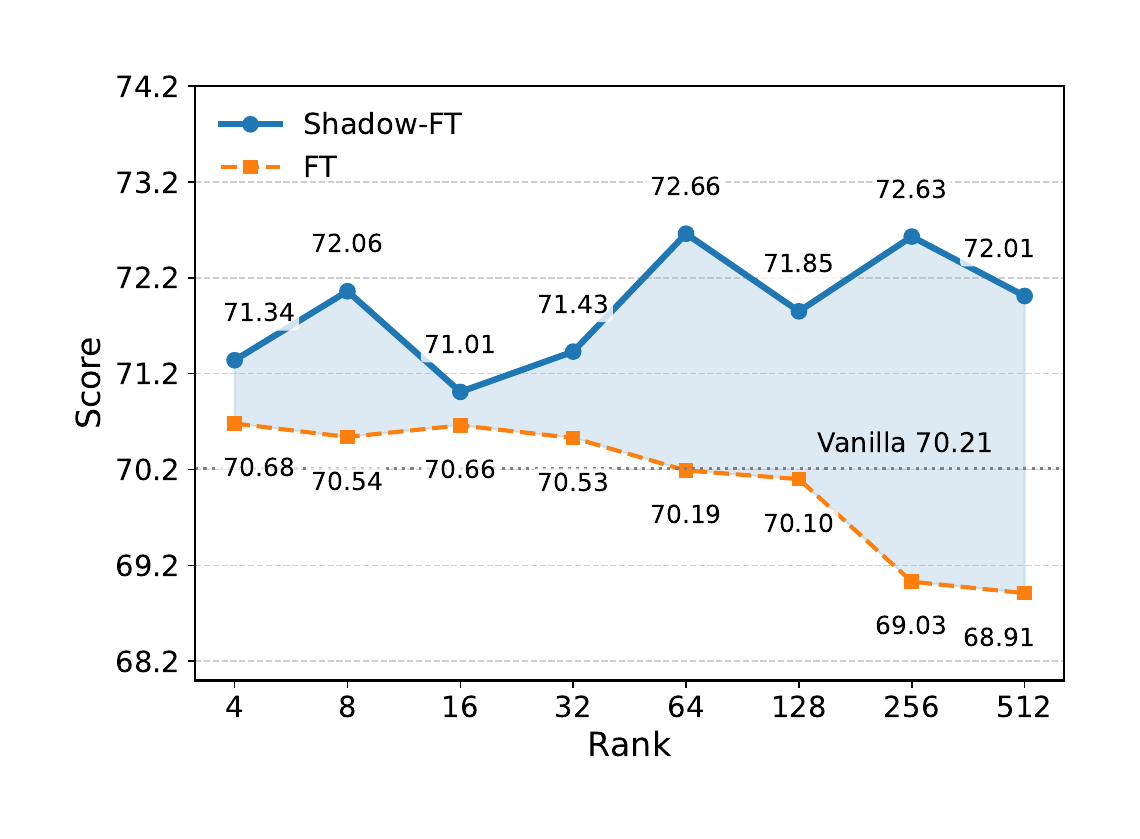}
\caption{
Performance tuning Llama-3.1-8B with different ranks.
}
\label{fig: rank_8b}
\end{figure*}

As shown in Figure \ref{fig: rank_8b}, the conclusions regarding Llama-3.1-8B are consistent with Section \ref{sec: ranks}.

\subsection{Weight Delta Scaling}
\label{appendix: weight_delta}

To further explore the effect of scaling the transferred deltas in the Shadow-FT strategy, 
we introduce an interpolation design controlled by a scaling factor $\alpha$. 
Specifically, let $\Delta W_B = W_B^{+} - W_B$ denote the weight updates learned on 
the \textsc{Base} model. Instead of directly applying the full deltas to the \textsc{Instruct} model, we interpolate as follows:
\begin{equation}
    W_{I}^{+} = W_I + \alpha \cdot \Delta W_B = W_I + \alpha \cdot (W_B^{+} - W_B).
\end{equation}

Here, $\alpha=1.0$ recovers the standard Shadow-FT formulation, while 
$\alpha=0.0$ reduces to the original \textsc{Instruct} model without transfer. 
Intermediate values of $\alpha$ provide a smooth interpolation between the two, 
allowing us to examine how the magnitude of transferred deltas affects 
downstream performance. The summarized results of Llama-3.1-8B-Instruct on the MATH-7 benchmark are 
presented in Table~\ref{tab:scaling}, showing that $\alpha=1.0$ yields the 
excellent overall trade-off, while other ratios offer insights into the sensitivity 
of tasks to partial transfer. Adaptive scaling strategies (e.g., layer-wise or task-specific factors) are left for future work.

\begin{table}[!h]
\centering
\caption{
Detailed results on the math benchmarks (averaged over three repeated runs for each ratio).
The ratio of \textbf{0.0} denotes \textbf{vanilla Instruct}, while \textbf{1.0} for the proposed \textbf{Shadow-FT}.
}
\label{tab:scaling}
\renewcommand{\arraystretch}{1.1}
{
\begin{tabular}{c|c|ccccccc}
\toprule
\textbf{Ratio} & \textbf{Math-7} & AIME24 & \makecell{GSM8K \\ (8-shot)} & \makecell{GSM8K \\ (0-shot)} & MATH & \makecell{MATH \\ 500} & \makecell{Minerva \\ Math} & SVAMP \\

\midrule
\rowcolor{green!15}
\makecell{0.0} & 56.2 & 1.1 & 84.8 & 85.0 & 48.5 & 50.7 & 40.5 & 82.7 \\ 
0.1  & 56.5 & 1.1 & 85.2 & 84.3 & 49.8 & 51.1 & 41.3 & 83.0 \\ 
0.2  & 57.1 & 2.2 & 85.0 & 85.0 & 50.4 & 50.5 & 42.3 & 84.3 \\ 
0.3  & 57.9 & 2.2 & 85.0 & 84.4 & 51.3 & 53.5 & 43.8 & 85.0 \\ 
0.4  & 58.8 & 6.7 & 85.6 & 83.6 & 52.9 & 54.2 & 44.2 & 84.7 \\ 
0.5  & 59.0 & 6.7 & 86.0 & 83.7 & 52.8 & 53.7 & 45.6 & 84.7 \\ 
0.6  & 59.1 & 4.4 & 86.1 & 83.5 & 52.8 & 54.7 & 47.2 & 84.7 \\ 
0.7  & 59.6 & 5.6 & 85.1 & 84.0 & 53.6 & 55.8 & 47.6 & 85.3 \\ 
0.8  & 59.4 & 5.6 & 84.9 & 84.5 & 53.1 & 53.6 & 48.8 & 85.3 \\ 
0.9  & 59.8 & 3.3 & 85.0 & 84.7 & 53.0 & 56.0 & 49.4 & 87.0 \\ 
\rowcolor{green!15}
\makecell{1.0} & 59.5 & 3.3 & 84.7 & 85.1 & 53.0 & 54.9 & 49.4 & 85.8 \\ 
1.2  & 59.1 & 0.0 & 84.8 & 86.0 & 52.6 & 56.6 & 49.1 & 84.3 \\ 
1.5  & 59.9 & 3.3 & 84.6 & 84.9 & 52.4 & 55.9 & 50.5 & 87.3 \\ 
2.0  & 57.9 & 0.0 & 83.2 & 84.2 & 50.2 & 52.0 & 48.1 & 87.7 \\ 
\bottomrule
\end{tabular}
}
\end{table}

\clearpage

\begin{table}[!h]
\centering
\scriptsize
\caption{
Detailed results on the math benchmarks for Table \ref{table: main_res}. Three times independent training and three times evaluation averages are reported.
}
\vspace{1em}
\label{appe_math}

\renewcommand{\arraystretch}{1.1}
\label{table: appendix_detail_math}
{
\begin{tabular}{cccccccccc}
\toprule

\textbf{Model} & \textbf{Method} & AIME24 & \makecell{GSM8K \\ (8-shot)} & \makecell{GSM8K \\ (0-shot)} & MATH & \makecell{MATH \\ 500} & \makecell{Minerva \\ Math} & SVAMP & \makecell{\textbf{Math-7}} \\
\midrule

\multirow{5}{*}{Llama-3.2-1B}  & Vanilla & 1.1 & 46.9 & 1.8 & 15.8 & 15.1 & 20.3 & 65.3 & 23.8 \\ 
 \cmidrule(lr){2-10}
 & FT (\textit{full}) & 1.1 & 46.8 & 0.8 & 18.6 & 19.1 & 19.0 & 66.5 & 24.5 \\ 
 & Shadow-FT (\textit{full}) & 0.0 & 47.2 & 1.0 & 18.9 & 18.3 & 23.1 & 67.9 & 25.2 \\ 
 \cmidrule(lr){2-10}
 & FT (\textit{LoRA}) & 1.1 & 45.2 & 2.6 & 21.8 & 20.3 & 18.7 & 67.5 & 25.3 \\ 
 & Shadow-FT (\textit{LoRA}) & 0.0 & 47.8 & 4.6 & 22.1 & 24.5 & 25.1 & 66.4 & 27.2 \\ 
\midrule
\multirow{5}{*}{Llama-3.2-3B}  & Vanilla & 4.4 & 76.6 & 79.5 & 45.8 & 47.9 & 36.1 & 84.7 & 53.6 \\ 
 \cmidrule(lr){2-10}
 & FT (\textit{full}) & 8.9 & 77.0 & 77.0 & 45.2 & 44.2 & 32.4 & 84.1 & 52.7 \\ 
 & Shadow-FT (\textit{full}) & 8.9 & 77.8 & 80.4 & 47.2 & 47.9 & 37.6 & 84.8 & 54.9 \\ 
 \cmidrule(lr){2-10}
 & FT (\textit{LoRA}) & 4.4 & 76.5 & 73.5 & 46.4 & 47.7 & 31.7 & 83.1 & 51.9 \\ 
 & Shadow-FT (\textit{LoRA}) & 11.1 & 78.1 & 77.6 & 49.8 & 52.0 & 39.3 & 85.4 & 56.2 \\ 
\midrule
\multirow{5}{*}{Llama-3.1-8B}  & Vanilla & 6.7 & 83.6 & 85.0 & 49.1 & 49.2 & 40.8 & 83.2 & 56.8 \\ 
 \cmidrule(lr){2-10}
 & FT (\textit{full}) & 1.1 & 84.0 & 85.4 & 51.0 & 51.5 & 39.0 & 85.8 & 56.8 \\ 
 & Shadow-FT (\textit{full}) & 6.7 & 85.0 & 84.0 & 52.2 & 53.2 & 43.8 & 86.3 & 58.7 \\ 
 \cmidrule(lr){2-10}
 & FT (\textit{LoRA}) & 6.7 & 83.8 & 83.8 & 50.2 & 52.5 & 41.3 & 86.5 & 57.8 \\ 
 & Shadow-FT (\textit{LoRA}) & 6.7 & 85.0 & 84.5 & 52.0 & 53.0 & 48.3 & 86.0 & 59.4 \\ 
\midrule
\multirow{5}{*}{Qwen-3-4B}  & Vanilla & 18.9 & 87.8 & 92.2 & 70.3 & 82.3 & 73.4 & 91.5 & 73.8 \\ 
 \cmidrule(lr){2-10}
 & FT (\textit{full}) & 14.4 & 88.1 & 91.6 & 70.1 & 82.4 & 72.1 & 91.2 & 72.9 \\ 
 & Shadow-FT (\textit{full}) & 16.7 & 87.4 & 92.3 & 70.0 & 84.3 & 73.3 & 91.7 & 73.7 \\ 
 \cmidrule(lr){2-10}
 & FT (\textit{LoRA}) & 18.9 & 84.2 & 91.6 & 68.1 & 77.3 & 67.4 & 90.7 & 71.2 \\ 
 & Shadow-FT (\textit{LoRA}) & 28.9 & 88.3 & 92.5 & 70.4 & 84.5 & 73.8 & 92.8 & 75.9 \\ 
\midrule
\multirow{5}{*}{Qwen-3-8B}  & Vanilla & 22.2 & 87.3 & 93.4 & 70.8 & 83.1 & 73.2 & 91.6 & 74.5 \\ 
 \cmidrule(lr){2-10}
 & FT (\textit{full}) & 22.2 & 86.2 & 93.1 & 70.6 & 80.7 & 72.7 & 92.1 & 74.0 \\ 
 & Shadow-FT (\textit{full}) & 32.2 & 87.5 & 93.3 & 70.6 & 82.9 & 73.2 & 91.4 & 75.9 \\ 
 \cmidrule(lr){2-10}
 & FT (\textit{LoRA}) & 17.8 & 83.6 & 92.1 & 68.9 & 77.3 & 68.4 & 90.7 & 71.3 \\ 
 & Shadow-FT (\textit{LoRA}) & 22.2 & 88.5 & 92.9 & 70.5 & 84.1 & 73.6 & 91.8 & 74.8 \\ 
\midrule
\multirow{5}{*}{Qwen-3-14B}  & Vanilla & 20.0 & 90.0 & 95.3 & 72.1 & 85.2 & 75.7 & 92.6 & 75.8 \\ 
 \cmidrule(lr){2-10}
 & FT (\textit{full}) & 17.8 & 88.9 & 94.9 & 72.2 & 85.5 & 75.5 & 91.3 & 75.1 \\ 
 & Shadow-FT (\textit{full}) & 40.0 & 90.7 & 95.2 & 71.7 & 86.3 & 76.0 & 92.7 & 78.9 \\ 
 \cmidrule(lr){2-10}
 & FT (\textit{LoRA}) & 14.4 & 87.3 & 94.5 & 71.7 & 81.3 & 72.8 & 91.0 & 73.3 \\ 
 & Shadow-FT (\textit{LoRA}) & 36.7 & 90.7 & 95.9 & 71.3 & 86.7 & 76.1 & 93.2 & 78.7 \\ 
\midrule
\multirow{5}{*}{Qwen-2.5-32B}  & Vanilla & 16.7 & 84.3 & 95.5 & 78.0 & 83.1 & 71.7 & 89.3 & 74.1 \\ 
 \cmidrule(lr){2-10}
 & FT (\textit{full}) & 21.1 & 86.6 & 95.4 & 74.8 & 82.9 & 76.8 & 92.1 & 75.7 \\ 
 & Shadow-FT (\textit{full}) & 13.3 & 85.0 & 95.5 & 76.8 & 84.1 & 78.0 & 91.3 & 74.9 \\ 
 \cmidrule(lr){2-10}
 & FT (\textit{LoRA}) & 14.4 & 85.7 & 95.3 & 73.6 & 83.8 & 75.0 & 92.1 & 74.3 \\ 
 & Shadow-FT (\textit{LoRA}) & 18.9 & 86.3 & 95.6 & 76.0 & 84.3 & 77.3 & 91.3 & 75.7 \\ 
\bottomrule

\end{tabular}
}
\end{table}

\begin{table}[ht]
\centering
\small
\caption{
Detailed data on the code benchmarks for Table \ref{table: main_res}. Three times independent training and three times evaluation averages are reported.
}
\vspace{1em}

\label{tab:appe_code}
\setlength{\tabcolsep}{4pt}
\renewcommand{\arraystretch}{1.15}
\begin{tabular}{llcccccccc}
\toprule

\textbf{Model} & \textbf{Method} & HumanEval & HumanEval$^{+}$ & \multicolumn{4}{c}{LiveCodeBench} & \textbf{Code-3}\\
 & & & & Exec & Gen & Out & Avg & \\
\midrule
\multirow{5}{*}{Llama 3.2-1B} 
  & Vanilla                 & 40.9 & 35.0 & 4.0 & 7.0 & 0.2 & 3.7 & 26.5\\
  \cmidrule(lr){2-9}
  & FT (\textit{full})      & 38.2 & 34.2 & 9.3 & 6.9 & 1.2 & 5.8 & 26.1\\
  & Shadow-FT (\textit{full})& 42.9 & 36.8 & 5.9 & 7.3 & 1.1 & 4.8 & 28.2\\
  \cmidrule(lr){2-9}
  & FT (\textit{LoRA})      & 39.2 & 34.4 &11.0 & 6.5 & 0.8 & 6.1 & 26.6\\
  & Shadow-FT (\textit{LoRA})& 41.9 & 35.4 &10.5 & 6.9 & 2.2 & 6.5 & 27.9\\
\midrule
\multirow{5}{*}{Llama 3.2-3B} 
  & Vanilla                 & 60.0 & 52.2 & 0.0 &16.8 & 0.7 & 5.8 & 39.3\\
  \cmidrule(lr){2-9}
  & FT (\textit{full})      & 57.9 & 53.7 & 4.5 &16.3 & 6.2 & 9.0 & 40.2\\
  & Shadow-FT (\textit{full})& 60.6 & 54.1 & 0.0 &16.8 & 2.1 & 6.3 & 40.3\\
  \cmidrule(lr){2-9}
  & FT (\textit{LoRA})      & 59.1 & 50.8 & 9.7 &17.1 &15.5 &14.1 & 41.4\\
  & Shadow-FT (\textit{LoRA})& 61.2 & 55.3 &14.4 &16.0 & 5.8 &12.1 & 42.9\\
\midrule
\multirow{5}{*}{Llama 3.1-8B} 
  & Vanilla                 & 69.7 & 62.8 &17.3 &19.8 &23.2 &20.1 & 50.9\\
  \cmidrule(lr){2-9}
  & FT (\textit{full})      & 70.7 & 67.3 & 16.9 &22.3 & 27.5 & 22.2 & 53.4\\
  & Shadow-FT (\textit{full})& 70.1 & 63.6 & 16.6 &20.8 & 27.8 & 21.7 & 51.8\\
  \cmidrule(lr){2-9}
  & FT (\textit{LoRA})      & 70.7 & 63.4 &16.3 &21.0 & 26.8 &21.4 & 51.8\\
  & Shadow-FT (\textit{LoRA})& 71.1 & 50.4 & 14.9 &21.3 &27.3 &21.2 & 50.9\\
\midrule
\multirow{5}{*}{Qwen-3-4B}   
  & Vanilla                 & 77.9 & 71.3 &41.8 &48.8 &59.7 &50.1 & 66.4\\
  \cmidrule(lr){2-9}
  & FT (\textit{full})      & 80.9 & 70.9 &43.1 &46.1 &53.0 &47.4 &66.4 \\
  & Shadow-FT (\textit{full}) & 80.3 & 71.1 &42.5 &49.7 &60.1 &50.8 & 67.4\\
  \cmidrule(lr){2-9}
  & FT (\textit{LoRA})      & 76.4 & 69.1 &13.1 &41.1 &45.6 &33.3 & 59.6\\
  & Shadow-FT (\textit{LoRA})& 81.3 & 76.8 &43.2 &49.1 &60.6 &51.0 & 69.7\\
\midrule

\multirow{5}{*}{Qwen-3-8B}   
  & Vanilla                 & 85.8 & 79.9 &42.3 &51.3 &63.4 &52.3 & 72.7\\
  \cmidrule(lr){2-9}
  & FT (\textit{full})      & 82.7 & 79.3 &42.9 &51.8 &59.7 &51.5 &71.2 \\
  & Shadow-FT (\textit{full})& 86.8 & 79.3 &41.9 &52.3 &65.2 &53.1 & 73.1\\
  \cmidrule(lr){2-9}
  & FT (\textit{LoRA})      & 84.2 & 78.5 &42.0 &45.7 &50.9 &46.2 & 69.6\\
  & Shadow-FT (\textit{LoRA})& 84.6 & 77.6 &41.9 &52.4 &66.1 &53.5 & 71.9\\
\midrule

\multirow{5}{*}{Qwen-3-14B}  
  & Vanilla                 & 86.8 & 83.1 &51.9 &55.8 &74.2 &60.6 & 76.8\\
  \cmidrule(lr){2-9}
  & FT (\textit{full})      & 87.6 & 83.5 &50.9 &54.3 &67.3 &57.5 & 76.2\\
  & Shadow-FT (\textit{full})& 87.4 & 82.9 &52.1 &55.6 &74.4 &60.7 & 77.0\\
  \cmidrule(lr){2-9}
  & FT (\textit{LoRA})      & 85.6 & 82.3 &51.2 &51.3 &62.5 &55.0 & 74.4\\
  & Shadow-FT (\textit{LoRA})& 87.8 & 84.4 &50.7 &56.8 &76.4 &61.3 & 77.8\\
\midrule
\multirow{5}{*}{Qwen-2.5-32B} 
  & Vanilla                 & 86.4 & 82.1 &58.3 &54.6 &64.6 &59.1 & 75.9\\
  \cmidrule(lr){2-9}
  & FT (\textit{full})      & 85.6 & 81.1 &60.3 &55.8 &66.4 &60.8 & 75.8\\
  & Shadow-FT (\textit{full})& 86.6 & 81.5 &60.5 &55.7 &64.0 &60.1 & 76.1\\
  \cmidrule(lr){2-9}
  & FT (\textit{LoRA})      & 85.4 & 81.7 &60.9 &55.0 &64.8 &60.7 & 75.9\\
  & Shadow-FT (\textit{LoRA})& 87.4 & 80.5 &61.8 &55.0 &64.9 &60.6 & 76.2\\
\bottomrule
\end{tabular}

\end{table}

\begin{table}[ht]
\centering
\scriptsize
\setlength{\tabcolsep}{4pt}
\caption{Detailed results on the general Reasoning benchmarks for Table~\ref{table: main_res}.}
\label{appendix_reasoning}

\renewcommand{\arraystretch}{1.15}
\begin{tabular}{lcccccccccc}
\toprule
\multirow{2}{*}{\textbf{Method}} & \multirow{2}{*}{MMLU} & \multicolumn{1}{c}{MMLU} & \multirow{2}{*}{WinoG} & 
\multirow{2}{*}{DROP} & \multicolumn{1}{c}{ARC} & \multicolumn{1}{c}{BBH} & 
\multicolumn{1}{c}{BBH} & \multicolumn{1}{c}{GPQA} & \multirow{2}{*}{TheoremQA} & 
\multirow{2}{*}{\textbf{Knowledge-9}} \\
 &  & Pro &  &  & Challenge & (0-shot) & (3-shot) & Diamond & & \\ 
\midrule
\addlinespace
 &  \multicolumn{9}{c}{\textit{Llama-3.2-1B}}\\
\addlinespace
\midrule
Vanilla                     & 46.8 & 21.4 & 51.9 & 42.7 & 56.6 & 24.4 & 26.1 & 27.8 &  9.9 & 34.2 \\
\cmidrule(lr){1-11}
FT (\textit{full})          & 46.9 & 21.8 & 50.4 & 39.0 & 56.6 & 20.4 & 24.8 & 24.8 & 10.8 & 32.8 \\
Shadow-FT (\textit{full})   & 47.1 & 22.7 & 51.1 & 41.6 & 52.9 & 22.2 & 20.8 & 26.3 &  9.6 & 32.7 \\
\cmidrule(lr){1-11}
FT (\textit{LoRA})          & 46.7 & 22.1 & 51.2 & 40.8 & 56.6 & 20.9 & 26.3 & 23.2 & 11.8 & 33.3 \\
Shadow-FT (\textit{LoRA})   & 46.6 & 23.2 & 51.4 & 43.9 & 52.2 & 17.2 & 20.4 & 25.3 & 10.6 & 32.3 \\
\midrule
\addlinespace
 &  \multicolumn{9}{c}{\textit{Llama-3.2-3B}}\\
\addlinespace
\midrule
Vanilla                     & 62.4 & 39.7 & 53.9 & 71.8 & 78.6 & 41.8 & 49.2 & 29.3 & 17.4 & 49.3 \\
\cmidrule(lr){1-11}
FT (\textit{full})          & 62.0 & 39.2 & 54.5 & 71.7 & 79.0 & 41.8 & 51.8 & 25.8 & 18.5 & 49.4 \\
Shadow-FT (\textit{full})   & 62.4 & 40.4 & 54.3 & 72.1 & 79.0 & 41.7 & 50.0 & 28.3 & 17.6 & 49.5 \\
\cmidrule(lr){1-11}
FT (\textit{LoRA})          & 61.9 & 39.9 & 51.1 & 71.7 & 82.7 & 41.4 & 49.4 & 25.3 & 18.4 & 49.1 \\
Shadow-FT (\textit{LoRA})   & 62.1 & 41.6 & 54.6 & 72.0 & 79.0 & 38.6 & 49.6 & 26.8 & 16.1 & 48.9 \\
\midrule
\addlinespace
 &  \multicolumn{9}{c}{\textit{Llama-3.1-8B}}\\
\addlinespace
\midrule
Vanilla                     & 69.5 & 48.5 & 59.4 & 81.4 & 85.4 & 44.6 & 67.6 & 25.8 & 27.3 & 56.6 \\
\cmidrule(lr){1-11}
FT (\textit{full})          & 69.7 & 49.2 & 60.9 & 80.0 & 87.1 & 46.8 & 71.1 & 30.8 & 30.6 & 58.5 \\
Shadow-FT (\textit{full})   & 69.6 & 49.3 & 60.2 & 81.7 & 85.8 & 46.8 & 67.0 & 28.3 & 29.5 & 57.6 \\
\cmidrule(lr){1-11}
FT (\textit{LoRA})  &69.3 & 48.9 & 60.0 & 79.5 & 86.4 & 48.8 & 68.0 & 30.3 & 26.8  & 57.5\\
Shadow-FT (\textit{LoRA}) & 69.4 & 50.8 & 60.2 & 80.1 & 85.4 & 51.6 & 68.8 & 32.8 & 29.1 & 58.7\\
 
\midrule
\addlinespace
 &  \multicolumn{9}{c}{\textit{Qwen-3-4B}}\\
\addlinespace
\midrule
Vanilla                     & 70.7 & 57.1 & 57.7 & 77.3 & 91.5 & 57.7 & 78.7 & 37.4 & 44.6 & 63.6 \\
\cmidrule(lr){1-11}
FT (\textit{full})          & 70.7 & 54.2 & 56.8 & 75.9 & 91.2 & 57.3 & 77.2 & 38.9 & 44.4 & 63.0 \\
Shadow-FT (\textit{full})   & 71.4 & 57.0 & 57.4 & 77.7 & 92.2 & 58.4 & 78.4 & 45.0 & 46.5 & 64.9 \\
\cmidrule(lr){1-11}
FT (\textit{LoRA})          & 71.9 & 51.2 & 59.0 & 69.1 & 91.5 & 54.7 & 73.5 & 39.4 & 39.9 & 61.1 \\
Shadow-FT (\textit{LoRA})   & 71.8 & 58.2 & 58.8 & 79.1 & 91.9 & 59.6 & 77.0 & 46.0 & 42.4 & 65.0 \\
\midrule
\addlinespace
 &  \multicolumn{9}{c}{\textit{Qwen-3-8B}}\\
\addlinespace
\midrule
Vanilla                     & 76.5 & 55.8 & 55.7 & 85.2 & 91.9 & 59.8 & 80.0 & 46.5 & 31.0 & 64.7 \\
\cmidrule(lr){1-11}
FT (\textit{full})          & 76.3 & 53.0 & 54.8 & 84.8 & 91.2 & 60.1 & 80.1 & 44.4 & 36.5 & 64.6 \\
Shadow-FT (\textit{full})   & 76.6 & 56.0 & 54.8 & 85.8 & 92.2 & 59.2 & 79.8 & 53.5 & 32.1 & 65.6 \\
\cmidrule(lr){1-11}
FT (\textit{LoRA})          & 76.1 & 57.2 & 55.9 & 80.6 & 92.5 & 59.0 & 75.5 & 41.4 & 40.9 & 64.3 \\
Shadow-FT (\textit{LoRA})   & 78.6 & 61.5 & 55.0 & 85.8 & 92.5 & 59.3 & 79.6 & 56.6 & 41.1 & 67.8 \\
\midrule
\addlinespace
 &  \multicolumn{9}{c}{\textit{Qwen-3-14B}}\\
\addlinespace
\midrule
Vanilla                     & 79.4 & 64.2 & 68.5 & 86.3 & 94.6 & 61.4 & 84.2 & 47.0 & 54.6 & 71.1 \\
\cmidrule(lr){1-11}
FT (\textit{full})          & 79.7 & 61.3 & 67.8 & 85.5 & 94.9 & 61.2 & 84.1 & 47.5 & 53.0 & 70.6 \\
Shadow-FT (\textit{full})   & 79.6 & 64.9 & 68.7 & 86.9 & 94.6 & 60.3 & 83.9 & 46.5 & 57.6 & 71.4 \\
\cmidrule(lr){1-11}
FT (\textit{LoRA})          & 79.6 & 60.7 & 68.5 & 84.0 & 95.3 & 63.3 & 83.0 & 47.0 & 51.9 & 70.4 \\
Shadow-FT (\textit{LoRA})   & 79.8 & 66.1 & 69.1 & 88.1 & 93.6 & 58.2 & 83.6 & 48.0 & 56.8 & 71.5 \\
\midrule
\addlinespace
 &  \multicolumn{9}{c}{\textit{Qwen-2.5-32B}}\\
\addlinespace
\midrule
Vanilla                     & 83.4 & 68.8 & 82.2 & 88.1 & 95.3 & 63.6 & 84.6 & 39.9 & 54.3 & 73.4 \\
\cmidrule(lr){1-11}
FT (\textit{full})          & 83.4 & 68.3 & 81.9 & 88.7 & 94.6 & 63.0 & 83.8 & 42.4 & 56.5 & 73.6 \\
Shadow-FT (\textit{full})   & 83.2 & 69.1 & 82.6 & 88.4 & 95.6 & 64.3 & 82.9 & 39.4 & 55.8 & 73.5 \\
\cmidrule(lr){1-11}
FT (\textit{LoRA})          & 83.6 & 68.9 & 82.2 & 88.8 & 94.9 & 62.8 & 83.7 & 44.4 & 54.8 & 73.8 \\
Shadow-FT (\textit{LoRA})   & 83.0 & 68.8 & 82.7 & 88.6 & 94.6 & 64.3 & 83.4 & 43.9 & 54.8 & 73.8 \\
\bottomrule
\end{tabular}
\end{table}

\clearpage

\section{Case Study on Llama-3.1-8B}
\label{app: case_study}

\begin{table}[]
\centering
\caption{Case study example from MATH-500. 
Shadow-FT can avoid \textit{errors} with the knowledge learned from the base model, while the vanilla tune method fails.
We extract the key step for better visualization.}
\label{tab:case-study}
\begin{tabular}{@{}l p{0.65\linewidth}@{}}
\toprule
\textbf{Input} &
\begin{minipage}[t]{\linewidth}\raggedright
\ttfamily
\detokenize{Question:} \\
\detokenize{How many positive two-digit integers are factors of both 100 and 150?}
\end{minipage}
\\
\midrule
\textbf{Ground Truth} &
\begin{minipage}[t]{\linewidth}\raggedright
\ttfamily
\detokenize{$\boxed{3}$ (The common two-digit factors are 10, 25, and 50).}
\end{minipage}
\\
\midrule
\textbf{Vanilla FT} &
\begin{minipage}[t]{\linewidth}\raggedright
\ttfamily
\verb|…| \\
\detokenize{1. **Find the GCD of 100 and 150:**} \\
\verb|…| \\
\detokenize{2. **Identify the two-digit factors of 50:**} \\
\verb|…| \detokenize{ stop counting at 25.} \\
\detokenize{Thus, the positive two-digit integers that are factors of both 100 and 150 are 10 and 25.} \\
\detokenize{$\boxed{2}$}
\end{minipage}
\\
\midrule
\textbf{Shadow-FT (Ours)} &
\begin{minipage}[t]{\linewidth}\raggedright
\ttfamily
\verb|…| \\
\detokenize{### Step 1: Find the Greatest Common Divisor (GCD) of 100 and 150} \\
\verb|…| \\
\detokenize{### Step 2: Find All Factors of the GCD (50)} \\
\verb|…| \\
\detokenize{### Step 3: Identify the Two-Digit Factors} \\
\verb|…| \\
\detokenize{(Note: 50 is a two-digit number too, but we'll check it as well.)} \\
\detokenize{So the two-digit factors of 50 are:} \\
\detokenize{$$10, 25, 50$$} \\
\detokenize{### Step 4: Count the Two-Digit Factors} \\
\verb|…| \\
\detokenize{### Final Answer:} \\
\detokenize{$$\boxed{3}$$}
\end{minipage}
\\
\bottomrule
\end{tabular}
\end{table}

Table \ref{tab:case-study} presents a representative case from Math-500 benchmark generated by Llama-3.1-8B-Instruct tuned via vanilla FT and Shadow-FT, respectively.
While the vanilla fine-tuned model partially solves the problem, it stops prematurely and misses one of the valid two-digit factors, resulting in an incorrect prediction of \detokenize{$\boxed{2}$}. 
In contrast, Shadow-FT correctly finds the GCD, enumerates all factors, and recognizes that 50 is also a two-digit factor, producing the correct answer \detokenize{$\boxed{3}$}. 
This example highlights the capability of Shadow-FT to leverage knowledge from the \textsc{Base} model and reason more comprehensively.

\section{Limitations}
In Shadow-FT, we first tune the \textsc{Base} model and then graft the weight updates to the \textsc{Instruct} model.
However, there are some LLMs for which the paired \textsc{Base} models are not available, such as Qwen3-32B and Qwen3-Next.
For these LLMs, we can not apply Shadow-FT.
Therefore, finding a proper 'shadow' for these models is an interesting topic for future work.

\end{document}